%% file: freeText_HC.tex
\documentclass[12pt]{article}

\input preambleHC.tex

\title{Machine Learning-Based Analysis of Free-Text Keystroke Dynamics}

\author{Han-Chih Chang\footnotemark[1]\ \ \ 
Jianwei Li\footnotemark[1]\ \ \ 
Mark Stamp\footnotemark[1]\,\,\footnotemark[2]}

\begin{document}

\symbolfootnotetext[1]{Department of Computer Science, San Jose State University}
\symbolfootnotetext[2]{mark.stamp$@$sjsu.edu}

\maketitle

\abstract
The development of active and passive biometric authentication 
and identification technology plays an increasingly important role in 
cybersecurity.  
Keystroke dynamics can be used to analyze the way that a user types 
based on various keyboard input. Previous work has shown that 
user authentication and classification can be achieved
based on keystroke dynamics. In this research, we consider the
problem of user classification based on keystroke dynamics
features collected from free-text.
We implement and analyze a novel a deep learning model that combines a 
convolutional neural network (CNN) and a gated recurrent unit (GRU).
We optimize the resulting model and consider several relevant related problems.
Our model is competitive with the best results obtained in previous 
comparable research.

\section{Introduction}

Recently, passive biometric authentication and identification techniques have 
received considerable attention. In this research, we consider such a 
passive biometric based on keystroke dynamics.
The resulting technique is applicable to the authentication problem, and can also
potentially play a role in intrusion detection.

Keystroke dynamics are based on typing behavior,
such as the duration of keyboard events, the duration of key presses, 
the time difference between key presses, and so on. 
Such data can be collected from a standard keyboard by monitoring 
input and recording the time intervals between each keystroke. 
Keystroke dynamics may not be strong enough
too be used as a standalone authentication system and hence it
is typically combined with another type of authentication, such as a password. 
In its role as an IDS, keystroke dynamics may be competitive 
with other techniques.

Compared with most other biometric technologies (e.g., fingerprint), 
keystroke dynamics has advantages. For example, no special hardware is required
to collect keystroke features, and
keystroke data can be obtained in a passive
manner, which reduces the burden on users. 
In addition, keystroke dynamics can be used as part of
a real-time intrusion detection system (IDS)---in contrast to a typical username 
and password authentication system, where ongoing
monitoring is not a realistic option. In addition to playing a role in authentication,
keystroke dynamics can serve to 
enhance security after a user has been authenticated. 

Disadvantages of keystroke dynamics based
authentication might arise if a user has an injured hand, a user is 
distracted, or the hardware (e.g., keyboard) has changed. 
We believe that that these---and related---disadvantages can be overcome, 
and we expect that the use of keystroke dynamics in security applications
will increase in the future.

In this paper, we analyze free-text keystroke dynamics data and train 
deep learning models to distinguish between users. 
The features we consider are related to keysytoke timing,
and are all obtained from standard keyboards. Furthermore,
we do not utilize the characters that are typed,
and hence user privacy is maintained.

We focus primarily on a novel model architecture that combines a
convolutional neural network (CNN) and a gated recurrent unit (GRU).
We experiment with non-timing features and we employ
concepts from Siamese networks~\cite{Siamese}, and we experiment 
with pre-trained models and attention.

Here, we only consider free-text data~\cite{bib3}. 
Specifically, for our free-text dataset, we use~\cite{9}, 
which contains keystroke data from~148 participants. 
The primary goal of this research is to achieve a high accuracy 
from a biometric system that uses features derived from keystroke dynamics. 
Among other desirable properties, we would ideally like our models to be 
scalable and robust. 

The following sections of this paper are organized as follows. Section~\ref{Background} 
discusses relevant background topics, with the focus on a survey of previous work. 
Section~\ref{Implementation} describes the dataset used in our experiments 
and outlines the machine learning techniques that we employ. 
Our  free-text experimental results are
presented in Section~\ref{Experiments and Results on Free-text Dataset},
while Section~\ref{Conclusion and Future Work} summarizes our 
main results, and include suggestions for future work.

\section{Background}\label{Background}

In this section, we discuss keystroke dynamics and we
provide a selective survey of previous work in this field. We also introduce the
machine learning models that we employ in this research. 

\subsection{Keystroke Dynamics}

According to~\cite{Monrose}, ``keystroke dynamics is not what you type, 
but how you type.'' Most previous work on typing biometrics can be 
divided into either classification that relies on fixed-text or 
authentication based on free-text data~\cite{4}. For fixed-text, the text used to 
model the typing behavior of a user and to authenticate the user is the same (e.g., a password). 
This approach is usually applied to short text sequences. 
Classification is generally based on timing features 
related to the character typed~\cite{10.1007/1-4020-8143-X_18}. 
An thorough discussion of the fixed-text problem 
can be found in~\cite{4}. 

For the free-text case, the text used to model typing 
behavior of a user and the text used to authenticate the same user is,
in general, not the same. 
Free-text usually implies long text sequences, and can be viewed as 
a continuous form of authentication or as part of an IDS.

In the past, distance based methods were popular for the
analysis of keystroke dynamics. More recently, machine learning techniques 
have come to the fore, including support vector machines (SVM), 
recurrent neural networks (RNN), hidden Markov models (HMM),
$k$-nearest neighbors ($k$-NN), multilayer perceptrons (MLP), and so on~\cite{Zhong}. 

Next, we discuss relevant previous work. Then, in the subsequent section,
we the dataset and learning techniques used in our experiments.

\subsection{Previous Work}

In this section, we first consider distance-based methods.
Then we discuss more recent work that is based on a wide
variety of (mostly) modern machine learning techniques.

\subsubsection{Distance Based Research}

The concept of keystroke dynamics can be traced back to the~1970s,
at which time the analysis was focused on fixed-text data~\cite{forsen1977personal}. 
Subsequently, Bayesian classifiers based on the mean and variance in 
time intervals between two or three consecutive key presses 
were applied to keystroke dynamics data~\cite{MONROSE2000351}.
For example, in~\cite{MONROSE2000351} the authors 
claim an accuracy of~92\%\ over a dataset consisting of~63 users. 

Typical of relatively early work in this field are
nearest neighbor classifiers based on distance measures.
Euclidean distance or, equivalently, the~$L_2$ norm was often used. 
More success was found using the~$L_1$ norm (i.e., Manhattan distance),
which makes it easier to single out the contribution made by individual
components, In addition, the~$L_1$ norm is more robust with respect to outliers. 
In~\cite{5}, the best result obtained from a distance-based technique uses 
a nearest neighbor classifier based on a scaled Manhattan distance.
Subsequently, statistical-based distance measures---such as 
Mahalanobis distance---were used with success
in keystroke dynamics research~\cite{62613}.

The equal error rate (EER) is a point on the ROC curve where the
sum of the false accept rate (FAR) and false rejection rate (FRR) is minimized. 
The EER is a commonly used measure of success for biometric 
systems---the lower the value of EER, the better the performance of 
a system. For example, in~\cite{5} the authors achieve an 
equal error rate (EER) of~0.096.

\subsubsection{Machine Learning Based Research}

Recently, research in keystroke dynamics has been dominated
by machine learning techniques. Such techniques 
include $K$-nearest neighbors ($k$-NN)~\cite{5634492}, 
$K$-means clustering~\cite{10.1007/978-3-540-74549-5_125}, 
random forests~\cite{5544311}, 
fuzzy logic~\cite{886039}, 
Gaussian mixture models~\cite{4656564}, 
and many, many more.

In comparison to the fixed-text problem, the number of research studies involving
free-text data is much smaller. In~\cite{122} it is claimed that
the amount of research done with fixed-text was eight times as much
as that for free-text, as of~2013. 

Free-text presents several challenges as compared to fixed-text.
For example, in free-text, the number of keys typed can differ greatly. 
There may also be word-specific dependencies in free-text~\cite{4270391}
that would not be relevant in fixed-text.

As an aside, we note that other keystroke features might be of interest.
For example, keystroke acoustics are considered 
in~\cite{6966780}, where a dataset containing~50 users 
yields an EER of~11\%. This results shows that acoustic-based typing information 
can be useful for authentication. However, an advantage of
keystroke dynamics based on timing features is that such information 
can be easily collected from any standard keyboard.

\section{Implementation}\label{Implementation}

In this section, we first introduce the keystroke dynamics dataset considered
in this research.
Then we discuss the various learning techniques that we have applied to this
free-text dataset.

\subsection{Dataset}

For our free-text data, we choose to 
use the so-called Buffalo dataset~\cite{9},
which was collected by researchers at SUNY Buffalo. This dataset
contains long fixed-text and free-text keystroke data from~157 subjects, with
each subject using the keyboard over three
 sessions~\cite{9}. For the fixed-text, 
users were requested to type Steve Jobs' Stanford commencement speech,
which was split into three pieces. In free-text, users are requested to 
answer two survey style questions and one scene description. 
The time duration within each session was about~50 minutes, 
with about~5700 keystrokes, on average, and hence over~17,000
for the three sessions combined. Furthermore, there was a~28 day 
time interval between sessions, and four different types of keyboards 
were used. In this paper, we only consider the Buffalo free-text keystroke data.

Note that the Buffalo free-text dataset is divided into two subsets, referred to as
a ``baseline'' subset and a ``rotation'' subset. In the baseline subset, there are~75 users 
using the same type of keyboard across all three sessions. 
For the rotation subset, there are~74 users using three different types 
of keyboard across their three sessions. The data collected on each keystroke
consists of the name of the key, the key event (key-down or key-up), 
and a timestamp (measured in milliseconds).

\subsection{Techniques Considered}

In our free-text experiments, we also employ machine learning models
that are somewhat more complex and experimental in nature, as compared to
those typically used in comparable research. Next, we briefly discuss 
both of the models we consider.

\subsubsection{BERT}

Bidirectional encoder representations from transformers (BERT) is a 
language model developed by 
Google~\cite{DBLP:journals/corr/abs-1810-04805}
that is designed to serve as an encoder in an
encoder-decoder model. 
The BERT encoder converts words into vector representations
that a corresponding decoder can then use to generate the output. 
BERT training is divided into two major steps which are 
known as pre-training and fine-tuning. In the pre-training phase, 
Google has used a large amount of text data to train the model in 
an unsupervised manner. In fine-tuning, labeled training data
is used to adapt the model to a specific problem. In our experiments,
we use the pre-trained model and the fine-tuning is performed based on
the words typed by a user.

\subsubsection{CNN-GRU Model}\label{CNNGRU}

We propose a novel hybrid CNN-GRU model
that is designed to learn from a sequence of individual keystroke features.
This model was inspired by related work in~\cite{10}. 

The core idea of using a GRU is that it can take advantage of sequential 
information in a user's typing behavior. Since a GRU is a type of 
recurrent neural network, it has the ability to learn the current characteristics 
of the input based on previous characteristics. In addition, we use a CNN 
before the GRU, with the aim of providing enhanced features to the GRU. 
In effect, this CNN step can be viewed as a form of feature engineering. 
In our CNN, the length of the convolutional kernel corresponds to the number 
of sequences that are covered. The convolution operation 
has the ability to produce a ``higher-level'' keystroke signature. 
Subsequently, these signatures serve as the input to the GRU. 
After training the GRU, a user's keystroke behavior pattern
is obtained.

We also implement dropout within the proposed model. 
The idea of dropout was introduced in~\cite{Srivastava} as a 
regularization technique for deep neural network.
As the name indicates, for dropouts we randomly drop nodes (neurons)
along with their connections from the neural network during training. 
This has the effect of training each mini-batch over a different network. 
Dropouts serve to prevent overfitting by forcing nodes
that would likely otherwise atrophy to be active in the learning process.

For this proposed model, we use the  \texttt{BCEWithLogitalLoss}
activation function, as opposed to the more typical sigmoid. 
According to~\cite{NEURIPS2019_9015}, \texttt{BCEWithLogitalLoss} 
is more stable than sigmoid or \texttt{BCELoss}.

\section{Free-Text Experiments}
\label{Experiments and Results on Free-text Dataset}

This section provides our experiments and results for the 
machine learning techniques discussed above when applied to 
the Buffalo free-text dataset. 
We also provide some analysis of our experiments. 

The Buffalo free-text dataset contains three different sessions.
Thus, we perform 3-fold cross validation with each session serving as
a fold. We use the notation s01-train-s2-test to mean that 
we use sessions~0 and~1 for training, with session~2 
reserved for testing---the notation for the other folds is analogous. 

\subsection{Text-Based Classification}\label{BERT}

First, we attempt to classify based on the text typed by users. 
Ultimately, we want to classify users based on their keystroke
dynamics, but by first focusing solely on the characters typed, we
can see how much information is contained in users' differing
responses, as opposed to their typing characteristics.

For our text-based experiment, we use BERT for multi-classification of the~148 
users in the free-text part of the Buffalo dataset, based on words typed. This experiment,
which we refer to as BERT-word,
was complicated by various typos that had to be corrected.

The result of this experiment is extremely poor, and testing loss does 
not decrease significantly. One problem may be that the dataset is too small,
as the training data for each user consists of only about~15 lines of words, 
with~2 lines for testing. Note that each line contains about~18 words.
In a typical NLP application, we would have hundreds of times
more data for training and testing.

The bottom line here is that we are not able to classify users based on
the actual text that was typed with an accuracy greater than guessing.
Although additional experiments might be helpful, it appears that
there is little useful information contained in the raw text. We now
turn our attention to keystroke dynamics based models.

\subsection{Keystroke Dynamics Models}

Here, we apply and analyze our novel~CNN-GRU architecture,
which we outlined in Section~\ref{CNNGRU}, above.
This model is based only on keystroke dynamics, as opposed to
the actual text typed by a user.

The goal here is to build a model that
could be used as part of an ongoing intrusion detection system (IDS). 
That is, the model would be used to periodically verify the identity of a
user in real time.

For our dataset, we employ the baseline subset of the Buffalo free-text,
in which each of~75 users typed across three sessions. After determining
the basic parameters of our model, we consider a wide variety of
modifications.

\subsubsection{Features}

We consider three types of features in our experiments,
namely, timing features, rate features (discussed below), 
and ``fusion.'' For the fusion case, we simply combine 
the timing and rate features.

First, we consider timing features only.
In this case, we transform the free-text keystroke data into a fixed-length 
keystroke sequence, then further convert the sequence into a keystroke 
vector. The format of the keystroke vector is presented 
in Figure~\ref{fig:vector}, where~$x$ and~$y$ are consecutive keystrokes. 
Note that~$x$ and~$y$ are simply numeric values 
representing the position in the keystroke sequence---the key
that was pressed in not recorded. That is, we do not use what was typed
as a feature, just how it was typed. This assures that a user's 
typing is not revealed by our analysis. 

The notation \texttt{H[$x$]} is the hold duration of the key,
\texttt{U[$x$]} is the key-up time (the timestamp when key is released),
and \texttt{D[$x$]} is the key-down time (the timestamp of the key was pressed). 
Therefore, \texttt{D[$x$]}\texttt{U[$x$]} is the time duration that a key 
was pressed until it was released, and \texttt{D[$x$]}\texttt{D[$y$]} 
is the time duration between two consecutive keys being pressed. 
In all cases, $x$ and~$y$ indicate any key being pressed.

\begin{figure}[!htb]
\centering 
\begin{tabular}{|c|c|c|c|c|c|}\hline
\texttt{ID[$x$]} & \texttt{ID[$y$]} & \texttt{H[$x$]} & \texttt{H[$y$]} 
	& \texttt{D[$x$]U[$y$]} & \texttt{D[$x$]D[$y$]}\\
\hline
\end{tabular}
\caption{Keystroke timing feature vector}
\label{fig:vector}
\end{figure}

After obtaining the keystroke vectors, we 
normalize the timing features. As discussed in~\cite{11},
this normalization results in features with mean~0 and variance~1.

Next, we consider ``rate'' features.
Traditionally, keystroke dynamics is based on timing, as discussed above. 
However, other typing habits can also serve as indicators of typing behavior. 
For example, features that relate to the use of the left and right hands
may help the model to distinguish between users.

In addition to conventional timing features, we further consider
seven features consisting of the rate
at which various keys are pressed. Specifically, we consider 
rate at which each of the delete, left shift, right shift, left caps, right caps, control key,
and the combination of left and right arrow keys are pressed.
These features will surely be useful for distinguishing a user's
handedness, but they may also be useful as more general
indications of typing style.

As a first experiment, we consider different numbers of key-strokes for analysis.
Specifically, we consider conventional key-length~100 and rate features 
with key-length~500. 
For each user, we append the rate 
features after the timing features. 
We have tested models with three different kinds of features, namely, 
timing only, rate only, and combined. 
The accuracy of the timing feature only is~84.72\%,
while the accuracy improves to~90.82\% after combining 
with the rate features.

\begin{figure}[!htb]
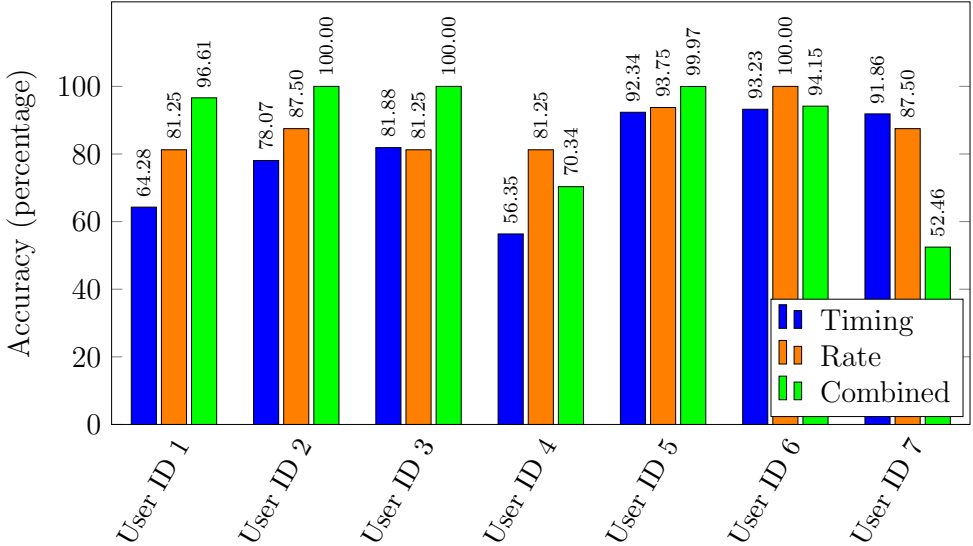

\centering
\input figures/barDiffUsers.tex
\caption{Accuracy for different users} 
\label{fig:adu}
\end{figure}

For some users, the rate features make a large difference,
but for others they actually make the results worse.
For example, user~1 with timing feature alone has an 
accuracy is only about~60\%. But if we combine with the
rate feature, accuracy increases to~95\%. 
In this case, the mixed features seem to be effective. 
However, for user~7, the timing feature alone yield an accuracy of~82\%,
but if we include the rate features, the accuracy
drops to~50\%\ for the fusion case, while considering rate alone, 
the accuracy is 87.5\%. In some cases it is better to use timing alone 
while in some other cases, it is better to combine both, and
yet other cases it is best to just use rate features. 
But in most cases, the accuracy is increased when using the combined features. 
The result of first seven users is shown in Figure~\ref{fig:adu}.

Finally, we consider feature ``fusion,'' that is, we combine timing 
and rate features, using the same key length for both. 
We have also experimented with independent key-lengths,
but this does not improve on the results presented here.

Next, we combine the rate features with the timing feature using the idea from 
Siamese network~\cite{Siamese}. A Siamese network has two inputs, which are fed 
into two neural networks that, respectively, map the inputs into a new space. 
A distance-related loss function is used to train the network parameters, 
so that the trained network can measure the similarity of the two inputs. 
For the rate features, we apply a linear transformation 
using the \texttt{torch.nn.Linear()} module from \texttt{PyTorch}, 
while conventional features use a multi-kernel CNN-GRU.
We then concatenate the two features together and pass 
them through a fully connected layer to obtain the desired output. 

The results of our fusion experiments are summarized in Table~\ref{tab:6}. 
We observe that when we apply longer key lengths, the result tend to improve. 
However, the reason we stop at key length~250 is that some users 
typed much less than was typically the case.
Regardless, these experiments show that the this fusion 
approach is effective, and achieves very high accuracy
for this authentication problem.

\begin{table}[!htb]
\caption{Feature fusion for different key lengths}\label{tab:6}
\centering
\adjustbox{scale=0.85}{
\begin{tabular}{c|cc|ccc|c}\midrule\midrule
\multirow{2}{*}{Model} & \multicolumn{2}{c|}{Parameters} & \multicolumn{3}{c|}{Test session} 
		& \multirow{2}{*}{Average} \\
	& kernel  & key-length & s0 & s1 & s2  \\ \midrule
\multirow{4}{*}{CNN-GRU} & 2,2,2  & 50   & 89.2\% & 89.7\% & 89.6\% & 89.5\% \\ 
     & 2,2,2  & 150  & 92.5\% & 93.1\% & 93.7\% & 93.1\% \\ 
     & 2,2,2  & 200  & 93.4\% & 93.6\% & 94.3\% & 93.8\% \\ 
     & 2,2,2  & 250  & 94.6\% & 94.1\% & 93.8\% & 94.2\% \\ 
\midrule\midrule
\end{tabular}
}
\end{table}

\subsubsection{Parameter Tuning}\label{pt}

Next, we perform parameter tuning on our CNN-GRU model. 
The hyperparameters that we vary include
the learning rate, kernel size of the CNN, and keystroke length,
among others. Since our model is implemented in Cuda~\cite{cuda},
we are able to efficiently test many parameter values.

As discussed above, our original model uses a single kernel CNN-GRU model 
and achieves an average accuracy of 84.7\%. 
We consider a multi-kernel CNN, where the kernel is a 
list so that different combinations can be tested. When we
include the rate features, 
we obtain a best average accuracy of~92.1\%\ using the model parameters 
in Table~\ref{tab:multi-kernel}.

\begin{table}[!htb]
\caption{Best result for parameter tuning of kernel}\label{tab:multi-kernel}
\centering
\adjustbox{scale=0.85}{
\begin{tabular}{c|ccc|c}\midrule\midrule
\multirow{2}{*}{Model} & \multicolumn{3}{c|}{Parameters} & \multirow{2}{*}{Accuracy} \\
 & kernel & out-channel & RNN-size \\ \midrule
CNN-GRU & 2,2,2 & 32 & 8 & 92.10\% \\
\midrule\midrule
\end{tabular}
}
\end{table}

In a multi-kernel model, when the kernel sizes differ, features 
are observed on different scales---when the kernel size is large,
the receptive field is bigger and more of the input is observed. 
The tradeoff is that larger kernel sizes tend to result in overfitting. 
Furthermore, when we combine different kernels together, 
padding issues arise. The purpose of padding is to make the size of the feature map 
consistent with the size of the original image, with padding determined by 
the size of the filter and the size of the stride. 
Padding enables us to use all of the actual data. 

Our multi-kernel experiment results are given in Table~\ref{tab:parametertuning}.
None of these results improve on our best previous accuracy of~92.3\%.

\begin{table}[!htb]
\caption{Parameter tuning CNN kernel}\label{tab:parametertuning}
\centering
\adjustbox{scale=0.8}{
\begin{tabular}{c|c|ccc|c}\midrule\midrule
\multirow{2}{*}{Model} & \multirow{2}{*}{Kernel} 
		& \multicolumn{3}{c|}{Test session} & \multirow{2}{*}{Average} \\ 
 &   & s0 & s1 & s2  \\ \midrule
\multirow{5}{*}{CNN-GRU} & 2        & 85.0\% & 85.6\%  & 83.5\% & 84.7\% \\ 
 & 2,4,6    & 89.2\% & 90.1\%  & 90.7\% & 90.0\% \\ 
 & 2,4,6,8  & 87.9\% & 88.1\%. & 87.9\% & 88.0\% \\ 
 & 2,2,2    & 91.7\% & 92.4\%. & 92.3\% & 92.1\% \\ 
 & 4,4,4    & 89.6\% & 90.4\%. & 90.0\% & 90.3\% \\ 
\midrule\midrule
\end{tabular}
}
\end{table}

Another parameter of interest in our CNN is the
``out channel'' which is the number of channels produced by the convolution.  
The results obtained when experimenting with
this parameter are given in Table~\ref{tab:outparametertuning}.
Here, we obtain a marginal improvement on our previous
best accuracy.

\begin{table}[!htb]
\caption{Parameter tuning CNN out channel}\label{tab:outparametertuning}
\centering
\adjustbox{scale=0.8}{
\begin{tabular}{c|cc|ccc|c}\midrule\midrule
\multirow{2}{*}{Model} & \multicolumn{2}{c|}{Parameters} 
	& \multicolumn{3}{c|}{Test session} & \multirow{2}{*}{Average} \\
 & kernel & out-channel  & s0 & s1 & s2  \\ \midrule
\multirow{6}{*}{CNN-GRU} & 2 & 16 & 90.9\% & 90.4\% &90.7\% & 90.6\% \\ 
 & 2 & 32 & 92.4\% & 92.1\% &91.4\% & 92.0\% \\ 
 & 2 & 48 & 92.6\% & 92.4\% &91.8\% & 92.3\% \\ 
 & 2 & 64 & 92.6\% & 92.2\% &91.8\% & 92.2\% \\ 
 & 2 & 96 & 91.6\% & 92.3\% &92.3\% & 92.1\% \\ 
 & 2 & 128 & 92.1\% & 92.5\% &91.6\% & 92.1\% \\
\midrule\midrule
\end{tabular}
}
\end{table}

We also conduct experiments varying the depths of convolutional layers.
These experimental results for three to nine layer modes 
are summarized in Table~\ref{tab:convparametertuning}. 
Somewhat surprisingly, these result indicate that higher layer models
do not seem to improve over our 3-layer model. 
In the realm of future work, it would be interesting to
experiment with other deep networks, 
such as ResNet, DenseNet, and SENet.

\begin{table}[!htb]
\caption{Parameter tuning CNN convolution}\label{tab:convparametertuning}
\centering
\adjustbox{scale=0.8}{
\begin{tabular}{c|cc|ccc|c}\midrule\midrule
\multirow{2}{*}{Model} & \multicolumn{2}{c|}{Parameters} 
	& \multicolumn{3}{c|}{Test session} & \multirow{2}{*}{Average} \\
 & Conv. & Learning rate  & s0 & s1 & s2  \\ \midrule
\multirow{4}{*}{CNN-GRU} & 3 & 0.001 & 91.1\% & 90.6\% & 89.9\% & 90.5\% \\ 
 & 3 & 0.01 & 91.6\% & 92.3\% & 91.6\% & 91.8\% \\ 
 & 6 & 0.01 & 92.1\% & 92.1\% & 91.8\% & 91.8\% \\ 
 & 9 & 0.01 & 91.5\% & 92.1\% & 90.8\% & 91.5\% \\
\midrule\midrule
\end{tabular}
}
\end{table}

Next, we experiment with an attention mechanism 
in both the single kernel and multi-kernel cases. 
When using conventional features only, we 
obtain a marginal improvement, as
summarized in Table~\ref{tab:attention}. 

\begin{table}[!htb]
\caption{Analysis of attention layer}\label{tab:attention}
\centering
\adjustbox{scale=0.8}{
\begin{tabular}{c|cc|ccc|c}\midrule\midrule
\multirow{2}{*}{Model} & \multicolumn{2}{c|}{Parameters} & \multicolumn{3}{c|}{Test session} & \multirow{2}{*}{Average} \\
 & kernel & attention  & s0 & s1 & s2  \\ \midrule
\multirow{4}{*}{CNN-GRU} & 2 & --- & 90.16\% & 88.37\% & 88.54\% & 89.02\% \\
 & 2 & \checkmark & 89.91\% & 89.17\% & 88.92\% & 89.00\% \\
 & 2,2,2 & --- & 91.70\%&92.40\%&92.30\%&92.10\% \\ 
 & 2,2,2 & \checkmark & 92.00\%&92.40\%&92.50\%&92.30\% \\
\midrule\midrule
\end{tabular}
}
\end{table}

After experimenting with additional parameter tuning, 
we find the best model uses the parameters in Table~\ref{tab:attention31}.
These parameters will be used in all subsequent experiments.

\begin{table}[!htb]
\caption{Attention and rate features (baseline subset)}
\label{tab:attention31}
\centering
\adjustbox{scale=0.85}{
\begin{tabular}{c|c|c|ccc|c}
\midrule\midrule
\multirow{2}{*}{Model} & \multirow{2}{*}{Hyperparameter} & \multirow{2}{*}{Value} 
	& \multicolumn{3}{c|}{Test session} & \multirow{2}{*}{Average}  \\ 
 & & & s0 & s1 & s2  \\ \midrule
\multirow{10}{*}{CNN-GRU} & CNN kernel & 2 & \multirow{10}{*}{95.4\%} 
	& \multirow{10}{*}{94.5\%} & \multirow{10}{*}{94.0\%} & \multirow{10}{*}{94.6\%} \\ 
& CNN out & 48 & & &   \\
& RNN size & 8 & & &   \\
& learning rate  & 0.001 &  &  & \\
& weight-decay  & 0.00001 & & &  \\
& step-scheduler  & 70 & & &  \\
& key-length & 250 & & &  \\
& epochs & 80 & & & \\
& attention & yes & & &  \\
& rate features & 7 & & &  \\
\midrule\midrule
\end{tabular}
}
\end{table}

\subsubsection{Fine Tuning}\label{fine-tune}

In this section, we consider a model for multi-classification of all~75 users
(as discussed above) and then use this model in a pre-trained mode to 
construct a binary classification model. We refer to this two-step
process as ``fine tuning.'' 

Note that here we consider binary classification of each user, 
and hence each user will have their own model. 
We then consider the average case for each of the resulting~75 models to obtain 
our accuracy results. 
In the multi-classification stage, we construct a single
model among~75 users that we then use as a pre-trained model
to construct each of the binary classifiers via ``fine-tuning.''

We have experimented with a wide variety of parameters at the multiclass
stage. We obtain a best result of~76.96\%\ using the parameters 
shown in Table~\ref{tab:multi-fine-final-1}.

\begin{table}[!htb]
\caption{Best parameters for multi-classification (baseline subset)}
\label{tab:multi-fine-final-1}
\centering
\adjustbox{scale=0.85}{
\begin{tabular}{c|c|c|ccc}
\midrule\midrule
\multirow{2}{*}{Model} & \multirow{2}{*}{Hyperparameter} 
	& \multirow{2}{*}{Value} & \multicolumn{3}{c}{Test session}  \\ 
 & & & s0 & s1 & s2  \\ \midrule
\multirow{8}{*}{CNN-GRU} & CNN kernel-size & 3 
	& \multirow{8}{*}{77.74\%} & \multirow{8}{*}{77.38\%} & \multirow{8}{*}{76.96\%} \\
& CNN out-channel & 192 & & &   \\
& RNN size & 32 & & &   \\
& learning rate  & 0.01 &  &  & \\
& weight-decay  & 1e-5 & & &  \\
& step-scheduler  & [80,350,390] & & &  \\
& key-length & 250 & & &  \\
& epochs & 400 & & & \\
\midrule\midrule
\end{tabular}
}
\end{table}

After obtaining our best pre-trained model, we further apply it in a binary 
classification model to obtain a best accuracy is~97.2\%, 
as summarized in Table~\ref{tab:fine-tune}. 

\begin{table}[!htb]
\caption{Fine-tune results (baseline subset with freeze)}\label{tab:fine-tune}
\centering
\adjustbox{scale=0.85}{
\begin{tabular}{c|cc|ccc|c}\midrule\midrule
\multirow{2}{*}{Model} & \multicolumn{2}{c|}{Parameters} 
	& \multicolumn{3}{c|}{Test session} & \multirow{2}{*}{Average} \\
 	& freeze & learning rate  & s0 & s1 & s2  \\ \midrule
\multirow{6}{*}{Fine-tune} & --- & 0.01\z\z    & 96.7\% & 97.4\% & 97.1\% & 97.1\% \\
 & --- & 0.001\z   & 97.3\% & 97.3\% & 96.9\% & 97.2\% \\
 & --- & 0.0001  & 94.5\% & 95.2\% & 94.6\% & 94.7\% \\
 & \checkmark  & 0.01\z\z    & 94.4\% & 94.7\% & 93.6\% & 94.2\% \\
 & \checkmark & 0.001\z   & 94.0\% & 93.9\% & 94.0\% & 94.0\% \\
 & \checkmark  & 0.0001  & 84.5\% & 84.5\% & 85.1\% & 84.7\% \\
\midrule\midrule
\end{tabular}
}
\end{table}

It can be seen from the result in 
Table~\ref{tab:ratio} that the ratio of positive samples to negative samples 
in the validation and test sets is roughly equal, which means that 
we have data with similar distributions for 
evaluation and testing. In Table~\ref{tab:ratio}, we 
consider user IDs from~1 to~5.

\begin{table}[!htb]
\caption{Test and validation ratios}\label{tab:ratio}
\centering
\adjustbox{scale=0.85}{
\begin{tabular}{c|ccc|ccc}\midrule\midrule
\multirow{2}{*}{User ID} & \multicolumn{3}{c|}{Test} & \multicolumn{3}{c}{Validation} \\
 & pos & neg & ratio & pos & neg & ratio\\ \midrule
1 & 310 & 12,797 & 0.02422 & 2643 & 115,319 & 0.0229 \\
2 & 182 & 12,925 & 0.01408 & 1673 & 116,289 & 0.0144 \\
3 & 248 & 12,859 & 0.01929 & 2109 & 115,853 & 0.0182 \\
4 & 173 & 12,934 & 0.01338 & 1697 & 116,265 & 0.0146 \\
5 & 184 & 12,923 & 0.01424 & 1607 & 116,355 & 0.0138 \\
\midrule\midrule
\end{tabular}
}
\end{table}

In addition, both the training and validation sets use random sampling, 
while the test set runs on all samples. 
Therefore, in the test set, the number of samples for label~0 (not users) 
is much more than the number of samples for label~1 (users). 
The results shown in Table~\ref{tab:re-run} indicate that a model 
that can better discriminate samples with label~0 will be stronger. 

\begin{table}[!htb]
\caption{Test and validation}\label{tab:re-run}
\centering
\adjustbox{scale=0.85}{
\begin{tabular}{c|ccc|c}\midrule\midrule
\multirow{2}{*}{Data} & \multicolumn{3}{c|}{Test session} & \multirow{2}{*}{Average} \\
  & s0 & s1 & s2  \\ \midrule
val   & 89.67\% & 87.98\% & 87.70\% & 88.45\% \\
test  & 94.74\% & 89.22\% & 90.05\% & 91.34\% \\
\midrule\midrule
\end{tabular}
}
\end{table}

We further analyze the precision, recall, F1 score, and perform 
parameter adjustments for the labels~0 and~1. 
Note that originally, label~1 is used as the positive label (user). 
However, since the data is so imbalanced, 
we swap the positive and negative label to observe the result on 
the multi-kernel~(2-2-2) model. These label switching results are
shown in Table~\ref{tab:f1}. We observe that
this model has a strong ability to detect intruders---the precision 
in the best case is virtually~100\%. This experiment indicates that our
model would work well as an IDS.

\begin{table}[!htb]
\caption{Label-swap}\label{tab:f1}
\centering
\adjustbox{scale=0.85}{
\begin{tabular}{c|ccc|c}\midrule\midrule
\multirow{2}{*}{Metric} & \multicolumn{3}{c|}{Test session} & \multirow{2}{*}{Average} \\
  & s0 & s1 & s2  \\ \midrule
Accuracy    & 94.74\% & 89.22\% & 90.05\% & 91.34\% \\
Precision   & 99.60\% & 99.58\% & 99.49\% & 99.57\% \\
Recall      & 95.05\% & 89.45\% & 90.40\% & 91.63\% \\
F1 Score    & 97.25\% & 94.04\% & 94.46\% & 95.25\% \\
\midrule\midrule
\end{tabular}
}
\end{table}

\subsubsection{GRU with Word Embedding}

In addition to keystroke features, we experiment with 
some text-base features in out GRU model. 
This does raise security concerns, since we must record what 
a user actually types, as opposed to simply using keystroke dynamics.
But, we want to determine whether this additional level of detail
can result in an improved model.

We use the \texttt{nn.Embedding} for word vectors in \texttt{PyTorch}, 
which provides a mapping between words and their corresponding vectors. 
The embedding weights can be trained, either by random initialization or 
by pre-trained word vector initialization. 
This technique can be used to determine
the positional relationship of two keys on the 
keyboard. For example, keys that are positioned next to each other 
can be classified as adjacent, and their vector should be similar. 
Furthermore, other relationships can be determine, such as 
keys that are pressed with the left hand, as compared to those that
are pressed with the right hand.

\begin{figure}[!htb]
\centering
\includegraphics[width=0.55\textwidth]{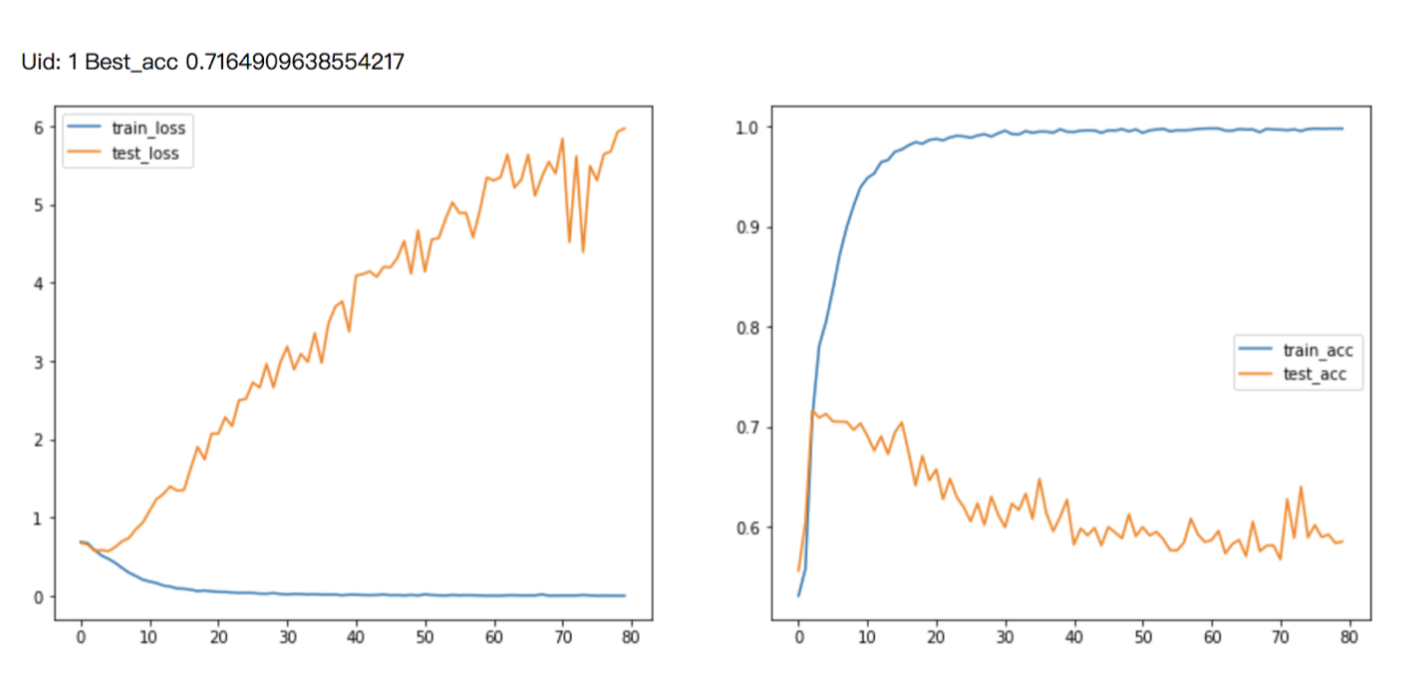}
\caption{Word embedding with GRU showing overfitting} 
\label{fig:word-gru-over}
\end{figure}

We implement this word embedding 
in our GRU model. Note that no CNN model is used in this experiment. 
We compare the experimental results for
different dimensions of word embedding vectors with and without attention. 
The result of these experiments are summarized 
in Table~\ref{tab:word_gru}. 
Note that embedding weights are trained by random initialization. 
Unfortunately, this model suffers from overfitting,
as can be seen from the graphs in Figure~\ref{fig:word-gru-over}.

\begin{table}[!htb]
\caption{Word embedding for GRU}\label{tab:word_gru}
\centering
\adjustbox{scale=0.85}{
\begin{tabular}{c|cc|ccc|c}\midrule\midrule
\multirow{2}{*}{Model} & \multicolumn{2}{c|}{Parameters} & \multicolumn{3}{c|}{Test session} & \multirow{2}{*}{Average} \\
 & embedding size & attention & s0 & s1 & s2  \\ \midrule
\multirow{6}{*}{GRU} & 3 & --- & 85.53\% & 83.68\% & 83.54\% & 84.25\% \\
 & 3 & \checkmark & 85.10\% & 84.52\% & 82.67\% & 84.10\% \\
 & 5 & --- & 83.22\% & 81.98\% & 80.76\% & 81.99\% \\
 & 5 & \checkmark & 83.41\% & 81.95\% & 80.64\% & 82.00\% \\
 & 10 & --- & 81.31\% & 79.29\% & 78.57\% & 79.72\% \\
 & 10 & \checkmark & 80.96\% & 79.27\% & 77.99\% & 79.41\% \\ \midrule\midrule
\end{tabular}
}
\end{table}

In an attempt to deal with this overfitting issue, we use
Word2Vec~\cite{w2v1,w2v2} to generate vector embeddings.
Specifically, we train on a sentence 
in two directions with random text length from~6 to~12 characters. 
We find that keys that are close together result in a higher score.
For example, \texttt{A} and~\texttt{S}, which are adjacent on a standard
QWERTY keyboard, have a cosine similarity of~0.9918, while~\texttt{A} 
and~\texttt{P}, which are on opposite ends of the keyboard, have
a cosine similarity of~0.7891.

The experimental results obtained of comparing random initialization and 
initialization with pre-trained vectors using a GRU model are
shown in Table~\ref{tab:word_pre_trained}.
Again, these results still clearly result in overfitting.
When embedding is used, the model can reach a training accuracy of 
about~0.998, but the loss during testing increases. 

\begin{table}[!htb]
\caption{Word embedding and pre-trained vectors}\label{tab:word_pre_trained}
\centering
\adjustbox{scale=0.85}{
\begin{tabular}{c|cc|ccc|c}\midrule\midrule
\multirow{2}{*}{Model} & \multicolumn{2}{c|}{Parameters} & \multicolumn{3}{c|}{Test session} & \multirow{2}{*}{Average} \\
 & embedding size & pre-trained & s0 & s1 & s2  \\ \midrule
\multirow{2}{*}{GRU} & 3 & --- & 85.53\% & 83.68\% & 83.54\% & 84.25\% \\
 & 3 & \checkmark & 84.99\% & 83.47\% & 82.53\% & 83.66\% \\
\midrule\midrule
\end{tabular}
}
\end{table}

When we apply word embedding in the CNN-GRU model, 
the overfitting issue is resolved---these results are given 
in Table~\ref{tab:improved-word-pre-trained}. However,
the result of word embedding do not outperform our previous experiment. 
This result is significant, since it shows that for our free-text dataset,
there is nothing to be gained by using the actual text typed by a user, 
as compared to simply using keystroke dynamics. Since using
the text would raise serious privacy concerns, it is beneficial that we
do not have to use such data to obtain optimal results.

\begin{table}[!htb]
\caption{Word embedding for CNN-GRU}\label{tab:improved-word-pre-trained}
\centering
\adjustbox{scale=0.85}{
\begin{tabular}{c|cc|ccc|c}\midrule\midrule
\multirow{2}{*}{Model} & \multicolumn{2}{c|}{Parameters} & \multicolumn{3}{c|}{Test session} & \multirow{2}{*}{Average} \\
 & kernel & embedding & s0 & s1 & s2  \\ \midrule
Word-CNN-GRU & 2 & 2 & 91.28\% & 93.07\% & 88.56\% & 90.07\% \\
CNN-GRU & 2,2,2 & None & 94.74\% & 89.22\% & 90.05\% & 91.34\% \\
\midrule\midrule
\end{tabular}
}
\end{table}

\subsubsection{CNN-Transformer}

Next, we experiment with a transformer technique
on our CNN-GRU model, which we refer to as 
the CNN-Transformer model. 
Specifically, we apply positional encoding before the 
encoder layer---the result of this experiment
is shown in Table~\ref{tab:transformer}.
This result is significantly worse than our previous best model,
so we do not pursue this approach further.

\begin{table}[!htb]
\caption{CNN-transformer results}\label{tab:transformer}
\centering
\adjustbox{scale=0.85}{
\begin{tabular}{c|ccc|c}\midrule\midrule
\multirow{2}{*}{Model} & \multicolumn{3}{c|}{Test session} & \multirow{2}{*}{Average} \\
  & s0 & s1 & s2  \\ \midrule
CNN-Encoder & 89.89\% & 86.22\% & 84.97\% & 87.03\% \\
\midrule\midrule
\end{tabular}
}
\end{table}

\subsubsection{CNN-GRU-Cross-Entropy-Loss}

Cross entropy can be used to determine how close the 
actual output is to the expected output. 
This loss function combines the two functions of \texttt{nn.LogSoftmax()} 
and \texttt{nn.NLLLoss()}. This function is considered useful when 
dealing with an imbalanced training set, and we have such a dataset.

In this experiment, we change the activation function from \texttt{BCEWithLogitsLoss} 
to \texttt{CrossEntropyLoss}. Furthermore, we calculate the output
during training and testing with \texttt{softmax} and \texttt{argmax}, respectively. 
The result for this experiment is given in Table~\ref{tab:cel}.
We see that this is a strong model, indicating that imbalance
may be an issue that we should address. Nevertheless, this experiment
does not improve on our best results.

\begin{table}[!htb]
\caption{CNN-GRU-cross-entropy results}\label{tab:cel}
\centering
\adjustbox{scale=0.85}{
\begin{tabular}{c|ccc|c}\midrule\midrule
\multirow{2}{*}{Model} & \multicolumn{3}{c|}{Test session} & \multirow{2}{*}{Average} \\
  & s0 & s1 & s2  \\ \midrule
CNN-GRU-Cross-Entropy & 98.3\% & 96.7\% & 95.2\% & 96.7\% \\
\midrule\midrule
\end{tabular}
}
\end{table}

\subsubsection{Rotation Subset}

To this point, our best model uses the fine-tuning technique discussed
in Section~\ref{fine-tune}. Based on this model, we 
also consider the so-called rotation subsets of the Buffalo free-text dataset, 
in which~73 users use different keyboards in different sessions. 

In this experiment, we build a multi-classification model on the rotation 
subset and obtain the result shown in Table~\ref{tab:keyboard}. Compared 
to our previous multi-classification results, the classification 
here is much worse. This is to be expected, since each session 
of the data is obtained from a different keyboard. 

\begin{table}[!htb]
\caption{Multi-classification on rotation subset}\label{tab:keyboard}
\centering
\adjustbox{scale=0.85}{
\begin{tabular}{c|cccc|c}\midrule\midrule
\multirow{2}{*}{Model} & \multicolumn{4}{c|}{Parameters} & \multirow{2}{*}{Accuracy} \\
 & kernel & out-channel & RNN-size & epochs \\ \midrule
\multirow{2}{*}{CNN-GRU} & 2 & 96 & 8 & 120,240 & 58.22\% \\
 & 16 & 192 & 64 & 40,80 & 49.16\% \\
\midrule\midrule
\end{tabular}
}
\end{table}

Next, we use a multi-classification model as a pre-trained model to 
build binary classifiers, analogous to the fine-tuned models discussed above.
The fine-tune results based on the rotation subset is shown in Table~\ref{tab:rotation}.

\begin{table}[!htb]
\caption{Fine-tuning on rotation subset (without freeze)}\label{tab:rotation}
\centering
\adjustbox{scale=0.85}{
\begin{tabular}{c|c|ccc|c}\midrule\midrule
\multirow{2}{*}{Model} & \multirow{2}{*}{Learning rate} & \multicolumn{3}{c|}{Test session} & \multirow{2}{*}{Average} \\
  & & s0 & s1 & s2  \\ \midrule
\multirow{2}{*}{CNN-GRU} & 0.001 & 86.9\% & 83.7\% & 91.1\% & 87.2\% \\
 & 0.01 & 89.8\% & 86.7\% & 93.2\% & 89.9\% \\
\midrule\midrule
\end{tabular}
}
\end{table}

From these results, we observe that the use of
different keyboards will likely create serious
difficulties for modeling based on keystroke dynamics. 
Thus, we conclude that different models will be needed
for the same user when using different keyboards.

\subsubsection{Robustness}

We also want to consider the robustness of our models. 
There are various definitions of robustness, but in general, 
we want to quantify the effect of a changing environment on a model.
There is no standard way to measure robustness for keystroke dynamics.
Here, we
use a technique known as synthetic minority oversampling technique (SMOTE) 
to generate synthetic data that
has similar characteristics to the training data. We then measure robustness
in the sense of how well our models perform on this SMOTE-generated data.

\begin{figure}[!htb]
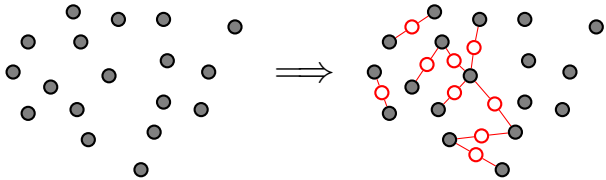

\centering
\begin{tabular}{ccc}
\input figures/smote1.tex
& 
\raisebox{0.5in}{\large$\Longrightarrow$}
& 
\input figures/smote3.tex
\end{tabular}
\caption{SMOTE illustrated}\label{fig:SMOTE}
\end{figure}

SMOTE is typically applied as a data augmentation technique to an imbalanced dataset. 
The idea behind SMOTE is to generate similar samples to the training
data using a straightforward interpolation approach~\cite{ger2020autoencoders}. 
The concept of SMOTE is illustrated in Figure~\ref{fig:SMOTE}, 
where the hollow circles on the right-hand side 
represent augmented data points that are obtained by interpolating
between actual data points. 

In this set of experiments, we use~$6\times 250$ dimensional array as the feature vector, 
and use the \texttt{imbalanced-learn} package in Python package 
to generate SMOTE data points.
First, we use SMOTE to increase the positive samples and apply a
``smoothing ratio'' of~0.1 which means that we increase in the number of 
positive samples to~0.1 times the number of negative samples. 
We also experiment with a smoothing ratio of~0.5. 
The training and validation results of these experiment are
summarized in Table~\ref{tab:s-val}.

\begin{table}[!htb]
\caption{SMOTE results for CNN-GRU}\label{tab:s-val}
\centering
\adjustbox{scale=0.85}{
\begin{tabular}{c|cc|c|ccc|c}\midrule\midrule
\multirow{2}{*}{Model} & \multicolumn{2}{c|}{Parameters} & \multirow{2}{*}{Result}
	& \multicolumn{3}{c|}{Test session} & \multirow{2}{*}{Average} \\
  & kernel & SMOTE ratio & & s0 & s1 & s2  \\ \midrule
\multirow{7}{*}{CNN-GRU}  & \multirow{2}{*}{2,2,2} & \multirow{2}{*}{---} 
 & Validation & 89.57\% & 87.56\% & 87.66\% & 88.26\% \\
	& & & Test & 94.74\% & 89.22\% & 90.05\% & 91.34\% \\ \cmidrule{2-8}
 & \multirow{2}{*}{8} & \multirow{2}{*}{0.1}  
 	& Validation & 79.76\% & 77.91\% & 77.93\% & 78.35\% \\
 	& & & Test &  98.05\% & 97.39\% & 96.50\% & 97.31\% \\ \cmidrule{2-8}
 & \multirow{2}{*}{8} & \multirow{2}{*}{0.5}  
 	& Validation & 73.47\% & 71.02\% & 71.32\% & 71.94\% \\
 	& & & Test &  98.49\% & 98.32\% & 98.07\% & 98.26\% \\
\midrule\midrule
\end{tabular}
}
\end{table}

The performance on the SMOTE
data shows that after augmenting, the accuracy decreases on 
the validation set, while the performance on the test set has improved. 
We speculate that SMOTE adds noise to the positive labels, which
worsens the ability of the model to judge positive cases. 
Moreover, when adding a higher proportion of SMOTE data, 
the performance on the validation set reduces further,
while performance on the test set improves. 

We also perform experiments for different ratios of under sampling. 
That is, we reduce the number of negative samples to
a specified proportion of the positive samples. 
The results of these experiments are shown in 
Table~\ref{tab:s-val-under}.
As expected, these results show minimal change, as compared
to the corresponding base models.

\begin{table}[!htb]
\caption{SMOTE results for CNN-GRU with undersampling}\label{tab:s-val-under}
\centering
\adjustbox{scale=0.85}{
\begin{tabular}{c|cc|c|ccc|c}\midrule\midrule
\multirow{2}{*}{Model} & \multicolumn{2}{c|}{Parameters} & \multirow{2}{*}{Result}
	& \multicolumn{3}{c|}{Test session} & \multirow{2}{*}{Average} \\
  & kernel & SMOTE ratio & & s0 & s1 & s2  \\ \midrule
\multirow{9}{*}{CNN-GRU}  & \multirow{2}{*}{2,2,2} & \multirow{2}{*}{---} 
 & Validation &  89.57\% & 87.56\% & 87.66\% & 88.26\% \\
	& & & Test &  94.74\% & 89.22\% & 90.05\% & 91.34\%  \\ \cmidrule{2-8}
 & \multirow{2}{*}{8} & \multirow{2}{*}{0.1 under}  
 	& Validation &  82.82\% & 81.40\% & 80.80\% & 81.67\% \\
 	& & & Test &  93.61\% & 89.90\% & 90.47\% & 91.33\%  \\ \cmidrule{2-8}
 & \multirow{2}{*}{8} & \multirow{2}{*}{0.5 under}  
 	& Validation & 82.70\% & 80.64\% & 80.49\% & 81.28\% \\ 
 	& & & Test &  93.38\% & 91.11\% & 89.96\% & 91.48\%  \\ \cmidrule{2-8}
 & \multirow{2}{*}{8} & \multirow{2}{*}{1.0 under}  
 	& Validation &  82.60\% & 81.09\% & 80.77\% & 81.49\% \\
 	& & & Test &  93.03\% & 91.17\% & 90.01\% & 91.40\%  \\
\midrule\midrule
\end{tabular}
}
\end{table}

We further analyze the precision, recall and F1 score of our models. 
Note that in these experiments we also perform label switching.
As discussed above, label switching may provide a better indication 
of the utility of a model in the IDS case. The results 
of these experiments are given in Table~\ref{tab:s-1.0}. 

\begin{table}[!htb]
\caption{SMOTE ratio~0.5 with label-switching}\label{tab:s-1.0}
\centering
\adjustbox{scale=0.85}{
\begin{tabular}{c|ccc|c}\midrule\midrule
\multirow{2}{*}{Metric} & \multicolumn{3}{c|}{Test session} 
	& \multirow{2}{*}{Average} \\
  & s0 & s1 & s2  \\ \midrule
Accuracy 	& 98.49\% & 98.22\% & 98.08\% & 98.26\% \\
Precision & 98.81\% & 98.79\% & 98.81\% & 98.80\% \\
Recall      & 99.67\% & 99.42\% & 99.24\% & 99.44\% \\ 
F1 Score  & 99.24\% & 99.10\% & 99.02\% & 99.12\% \\
\midrule\midrule
\end{tabular}
}
\end{table}

Comparing the result for the ``label-switch'' case in Table~\ref{tab:f1}, 
we conclude that the higher the SMOTE ratio, the higher the recall, 
and the lower the precision. This implies that the model is more 
capable of capturing data with label~0 (which, due to label switching, 
represents the positive case), but the accuracy of the model's judging 
label as~0 is also lower. 

\subsubsection{Explainability}

Next, we briefly consider the ``explainability'' of our model. That is, we would like
to gain some insight into how the model is actually making decisions. Most machine
learning techniques are relatively opaque, in the sense that it is difficult to 
understand the decision-making process. This is especially true of 
neural network based techniques, and since our model combines 
multiple techniques, it is bound to be even harder it interpret directly.
 
Here, we consider local interpretable model-agnostic 
explanations (LIME)~\cite{lime} to try to gain insight into the role of the
various features in our model. LIME perturbs 
the input and compares the corresponding outputs. 
If a small change in an input feature causes the classification to switch,
then we can judge the importance of a specific feature to the
model's decision-making process.

\begin{figure}[!htb]
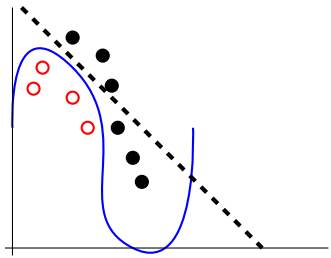

\centering
\input figures/lime.tex
\caption{LIME illustrated}\label{fig:lime}
\end{figure}

The concept behind LIME is illustrated in Figure~\ref{fig:lime},
where the solid black circles and hollow red circles represent two categories,
and the blue curve is the decision boundary between the classes.
LIME uses a simplified linear model---represented in the figure by
the dashed line---to predict the classes. Locally, this linear model will likely 
be very accurate, although it would typically fail badly globally.
By using a simple linear model, we can, for example, more easily determine
the most relevant features.

In our LIME experiment, we select user~46 and 
the ``s12-train-s0-test'' case. For this experiment, the best validation accuracy 
we obtain is 96.88\%\ and the best test accuracy is 99.31\%. 

For the sake of brevity, we omit the details of our LIME
experiments but, in summary, we find that LIME indicates that our 
model focuses more on holding time and difference time. 
In general, it appears that a good model focuses less on key-id, 
which indicates that it may not be a good feature to distinguish users.

\subsubsection{Equal Error Rate}

The equal error rate (EER) is an objective standard to measure classifiers. 
In a biometric system, the false accept rate (FAR) is the rate at which a user
can authenticate as someone else,
whereas the false reject rate (FRR) is the rate at which a user cannot
authenticate as themselves. When these rates are equal, 
the value is called equal error rate. The lower the EER value, 
the higher the accuracy of the biometrics system. 
In practice, we wold likely not set the system parameters so that the
FAR is equal to the FRR. For example, in a financial application 
that requires high security, the FAR should be very low, 
which necessitates a somewhat higher FRR. Nevertheless, the EER
allows us to easily compare different biometric systems.

Here, we use a sigmoid function for output,
so that we an obtain the prediction result in the form of
a probability. Then we use the prediction and ground-truth to calculate the confusion 
matrix. After obtaining labels, we construct an ROC curve 
to obtain the FPR and TPR and, finally, we determine the 
EER from the ROC curve. 

The EER results with s01 for training and s2 for testing 
for various models discussed above are provided in Table~\ref{tab:eer}. 
The best EER we obtain is~0.0386,
and when we use this model over all sessions we obtain 
an average EER of~0.0394.

\begin{table}[!htb]
\caption{EER for various models (baseline subset)}\label{tab:eer}
\centering
\adjustbox{scale=0.85}{
\begin{tabular}{l|c}\midrule\midrule
\multicolumn{1}{c}{Model} & s01-train-s2-test \\ \midrule
Pre-trained word embedding with CNN-GRU & 0.1091  \\
CNN-Transformer-encoder & 0.1257 \\
CNN-GRU-cross-entropy-loss & 0.1502  \\
CNN-GRU-without-sampler-at-best-val & 0.0609  \\
CNN-GRU-without-sampler-at-best-eer & 0.0611  \\
CNN-GRU-without-sampler-at-non-best-val & 0.0594   \\
CNN-GRU-with-sampler & 0.0826   \\
CNN-GRU-without-sampler-fine-tune & 0.0412 (0.7187-multi) \\
CNN-GRU-without-sampler-fine-tune & 0.0386 (0.7599-multi) \\
\midrule\midrule
\end{tabular}
}
\end{table}

After determining the best result on the baseline subset, we then apply this model to 
all~148 users. 
The resulting EER for all subset for different models is 
shown in Table~\ref{tab:eer-all}. However, we see that the result 
is poor, which is apparently due to the rotation 
subset, where different keyboards are used across sessions,
and the fact that some users do not 
type much data in the baseline subset. Thus, it is more
realistic to set a threshold for the key 
length with different users---in this case, 
we obtain a best EER result in Table~\ref{tab:eer-all-best}. 
From the last two lines in Table~\ref{tab:eer-all-best}, we 
observe that the model that has the higher multi-classification accuracy
results in a lower EER for the fine-tuned case.

\begin{table}[!htb]
\caption{EER for various models (all subsets)}\label{tab:eer-all}
\centering
\adjustbox{scale=0.85}{
\begin{tabular}{l|c}\midrule\midrule
\multicolumn{1}{c}{Model} & s01-train-s2-test \\ \midrule
CNN-GRU-attention-without-sampler & 0.1389  \\
CNN-GRU-attention-non-without-sampler & 0.1239  \\
CNN-GRU-fine-tune & 0.1029 \\
\midrule\midrule
\end{tabular}
}
\end{table}

\begin{table}[!htb]
\caption{Best model for EER (all subsets)}\label{tab:eer-all-best}
\centering
\adjustbox{scale=0.85}{
\begin{tabular}{c|ccc|c}\midrule\midrule
\multirow{2}{*}{Model} & \multicolumn{3}{c|}{Test session} & \multirow{2}{*}{Average} \\
  & s0 & s1 & s2  \\ \midrule
CNN-GRU-without-sampler-fine-tune & 0.0690 & 0.0841 & 0.0557 & 0.0696 \\
\midrule\midrule
\end{tabular}
}
\end{table}

\subsubsection{Knowledge Distilling}

Knowledge distilling is somewhat analogous to explainability.
In knowledge distillation we ``compress'' a model,
in the sense that we try to replace a complex model with
a much simpler one that achieves comparable results.
The goal is to extract the essence of a complex model
within a much simpler form.

This method knowledge distilling was first proposed 
in~\cite{10.1145/1150402.1150464}. Then in~\cite{hinton2015distilling}
a ``teacher'' and ``student'' model was proposed from the concept of mentoring. 
The output of the teacher network is used as a soft label to train a student network. 
For our model, the EER results obtained based on teacher-student 
knowledge distilling with different parameters are shown in Table~\ref{tab:kd}.

\begin{table}[!htb]
\caption{EER results for knowledge distilling}\label{tab:kd}
\centering
\adjustbox{scale=0.85}{
\begin{tabular}{l|c}\midrule\midrule
\multicolumn{1}{c}{Model} & s01-train-s2-test \\ \midrule 
Student-1-Teacher-1 & 0.1333  \\ 
Student-1.5-Teacher-0.5 & 0.1409  \\ 
Student-1.99-Teacher-0.01 & 0.1762 \\ 
Student-0.99-Teacher-0.01 & 0.1355 \\ 
Student-0.5-Teacher-1.5 & 0.0864 \\ 
Student-0.5-Teacher-1.99 & 0.1097 \\ 
Student-0.01-Teacher-1.99 & 0.0925  \\ 
\midrule\midrule
\end{tabular}
}
\end{table}

Note that in this experiment, we simply use a multi-classification model as the teacher
and the corresponding binary classification as the student model. With this type of approach, 
we hope to see that the teacher can improve the results given by 
the student model, as indicated by a low EER. However, in our experiments 
the EER is not competitive with our fine tuning model. Hence, we cannot
draw any strong conclusions from this experiment, but this is worth
pursuing as future work.

\subsubsection{Weighted Loss}

In the CNN-GRU fine-tune model, 
we select \texttt{BCEWithLogitsLoss} as our criterion to calculate the loss. 
However, we did not specify the parameter of \texttt{pos-weight} 
which is the weight of positive samples in the original experiment. 
In this experiment, we use the fine-tune model as a backbone and 
try to tune the positive weights and 
compare with our previous results.
The results of these experiments are given in Table~\ref{tab:wl}. 

\begin{table}[!htb]
\caption{Results for tuning \texttt{pos-weight} (baseline subset)}\label{tab:wl}
\centering
\adjustbox{scale=0.85}{
\begin{tabular}{c|c|cccc}\midrule\midrule
\multirow{2}{*}{Metric} & \multirow{2}{*}{Fine-tune} & \multicolumn{4}{c}{\texttt{pos-weight}} \\
  & & 0.1 & 2 & 10 & 50  \\ \midrule
ERR         & 0.0395  & 0.0428  & 0.0413  & 0.0392  & 0.0813 \\
Accuracy    & 99.33\% & 99.41\% & 99.21\% & 99.03\% & 96.81\% \\
Precision   & 78.46\% & 84.57\% & 74.82\% & 67.51\% & 50.87\% \\
Recall      & 76.88\% & 69.45\% & 78.92\% & 83.28\% & 82.77\% \\
F1 Score    & 74.63\% & 72.92\% & 73.40\% & 71.64\% & 56.64\% \\
AUC         & 98.84\% & 98.74\% & 98.87\% & 98.88\% & 98.28\% \\
\midrule\midrule
\end{tabular}
}
\end{table}

Note that the results in Table~\ref{tab:wl} are average among the three different test sessions. 
From the result in Table~\ref{tab:wl}, we observe that when the value of \texttt{pos-weight} is~0.1, 
the precision is best, while the recall rate decreases, as 
compared to the original fine-tune model. 
Moreover, when the value of \texttt{pos-weight} is higher, 
the precision decreases while the recall increases, which nicely illustrates the inherent 
trade-off between these measures. 
Based on these results, we can adjust the value of \texttt{pos-weight}
depending on different application scenarios. 

\subsubsection{Ensemble Models}

In this section, we consider three ensemble models which
include various combinations of fine-tuning, softmax, and transformer.
The results for these cases are given in Table~\ref{tab:ensemble}. 
Note that the results in Table~\ref{tab:ensemble} are each 
an average among three different test sessions. 
Since the results for the softmax and transformer models are much worse than the fine-tune model, 
the resulting ensembles only improve slightly in terms of precision and recall, and only for some users. 

\begin{table}[!htb]
\caption{Result for ensembles (baseline subset~1)}\label{tab:ensemble}
\centering
\adjustbox{scale=0.85}{
\begin{tabular}{c|c|cccc}\midrule\midrule
\multirow{2}{*}{Metrics} & \multirow{2}{*}{Ensemble} & \multicolumn{3}{c}{Model} \\
  & & fine-tune & softmax & transformers   \\ \midrule
ERR         & 0.082   & 0.0395  & 0.1644  & 0.1449    \\
Accuracy    & 99.28\% & 99.33\% & 96.80\% & 87.03\%   \\
Precision   & 76.71\% & 78.46\% & 41.69\% & 19.10\%   \\
Recall      & 78.34\% & 76.88\% & 71.30\% & 76.26\%   \\
F1 Score    & 74.29\% & 74.63\% & 47.46\% & 23.21\%   \\ 
AUC         & 95.18\% & 98.84\% & 84.23\% & 93.37\%   \\
\midrule\midrule
\end{tabular}
}
\end{table}

We further ensemble the models with different \texttt{pos-weight}
values, with the fine-tune model as the backbone. 
The parameters of the three ensembles considered, which we denote as
models~$\cal A$, $\cal B$, and~$\cal C$, are specified in Table~\ref{tab:comb}.
Note that the results in Table~\ref{tab:ensemble1} represent the 
average among the three different test sessions. 
We observe that the EER, precision, and recall rates all improve
with these ensemble techniques.

\begin{table}[!htb]
\caption{Various model combinations}\label{tab:comb}
\centering
\adjustbox{scale=0.85}{
\begin{tabular}{c|c}\midrule\midrule
Model & Combination \\ \midrule
$\cal A$ & $\mbox{fine-tune}, \mbox{pos-w~0.1}, \mbox{pos-w~2}, \mbox{pos-w~10}, \mbox{pos-w~50}$  \\
$\cal B$ & $\mbox{fine-tune}, \mbox{pos-w~0.1}, \mbox{pos-w~2}, \mbox{pos-w~10}$ \\
$\cal C$ & $\mbox{fine-tune}, \mbox{pos-w~0.1}, \mbox{pos-w~10}$ \\
\midrule\midrule
\end{tabular}
}
\end{table}

\begin{table}[!htb]
\caption{Results for ensembles (baseline subset~2)}\label{tab:ensemble1}
\centering
\adjustbox{scale=0.85}{
\begin{tabular}{c|c|cccc}\midrule\midrule
\multirow{2}{*}{Metrics} & \multirow{2}{*}{fine-tune} & \multicolumn{3}{c}{Model} \\
  & & $\cal A$ & $\cal B$ & $\cal C$   \\ \midrule
ERR         & 0.0395  & 0.0285   & 0.0312   &  0.0322  \\
Accuracy    & 99.33\% & 99.41\%  & 99.69\%  &  99.37\% \\
Precision   & 78.46\% & 81.31\%  & 82.98\%  &  82.32\% \\
Recall      & 76.88\% & 80.92\%  & 78.79\%  &  77.99\% \\
F1 Score    & 74.63\% & 78.67\%  & 78.31\%  &  77.75\% \\ 
AUC         & 98.84\% & 99.28\%  & 99.24\%  &  99.18\% \\
\midrule\midrule
\end{tabular}
}
\end{table}

\subsubsection{Discussion}

\begin{figure}[!htb]
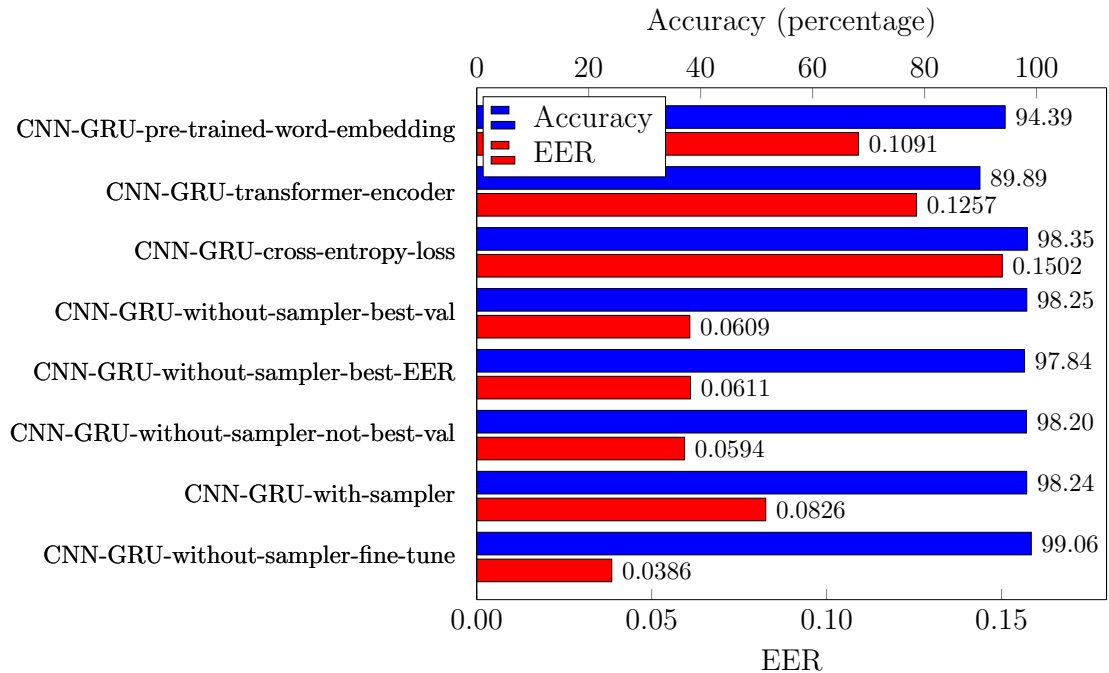

\centering
\input figures/barAccuracyPlusEER.tex
\caption{Accuracies and EER of models (baseline subset)} 
\label{fig:abeb}
\end{figure}

Figure~\ref{fig:abeb} summarizes the
results of our free-text experiments, both in terms of EER and accuracy.
Note that the best accuracy and the best EER were
both achieved with the CNN-GRU-without sampler-fine-tune model.

In our experiment, we have applied a sampler to train so as
to deal with the situation of imbalanced data. 
As for feature engineering, we transform the data into a vector which includes 
the label of the key and the timing features. Then, we compute the mean and 
variance and center the data, which enable us to achieve better performance. 

We perform parameter tuning on the models and obtain 
a great improvement in accuracy. The result shows that longer keystroke 
sequence and larger out-channel size generally result in higher accuracy, while more 
convolutions do not improve our model. 
After parameter tuning, we apply an attention layer on the outputs of 
the GRU and find that some users' accuracy slightly improves. Moreover, we 
expand the keystroke dynamics features to include rate features. 
Although the accuracy result of do not greatly improve, 
the EER does achieve better performance. 

We also use a multi-classification model as a pre-trained model for
binary classification. The results indicate that such a pre-trained 
multi-classification model can achieve higher accuracy 
and a lower EER for the corresponding binary classification model.

Finally, we compare our free-text experiments to previous 
work. From the results in Table~\ref{tab:pwr-free}, we see that our
best model is competitive with the best EER previously obtained,
while the accuracy of several of our models exceed~99\%.

\begin{table}[!htb]
\caption{Comparison to previous work for Buffalo free-text dataset}\label{tab:pwr-free}
\centering
\adjustbox{scale=0.85}{
\begin{tabular}{c|c|c|c}\midrule\midrule
Research & Models & Accuracy & EER \\ \midrule
Lu, et al.~\cite{10}      & CNN-RNN & --- & 0.0236\\
Ayotte, et al.~\cite{9123909} & ITAD metrics & --- & 0.0530 \\
Huang, et al.~\cite{8267670} & SVM & --- & 0.0493 \\ \midrule
Our research            & fine-tune CNN-GRU & Greater than 99\% & 0.0690\\
\midrule\midrule
\end{tabular}
}
\end{table}

\section{Conclusion and Future Work}
\label{Conclusion and Future Work}

In this paper, we developed and analyzed 
machine learning techniques for biometric 
authentication based on free-text data.
We focused on a novel CNN-GRU architecture, and we experimented
with an attention layer, rate features, pre-trained models, 
ensembles, and so on. 
The maximum multiclass classification accuracy that we achieved with 
our model was~99.31\%, with an EER of~0.069. As far as we are aware,
this is the best accuracy attained to date for the Buffalo free-text dataset,
and our EER is competitive with the best results previously obtained. 
In addition, for our CNN-GRU model, we considered ``explainability'' and 
knowledge distillation, among many other relevant topics.

The high accuracy and low EER for our free-text results indicate that 
passive authentication and intrusion detection may be practical,
based on keystroke dynamics. That is, in addition to an initial authentication
at login time, a user can be periodically re-authenticated by passively monitoring
typing behavior. In this way, intrusions can be detected in real-time,
with a minimal burden placed on users. 

There are many avenues available for future work. For example,
we plan to perform extensive model optimization and model fusion. 
For model optimization, we will consider
techniques from contrastive learning and self-supervised technique 
to see whether these approaches can improve our model. 

As another example of possible future work, we plan to evaluate the
robustness of our technique using an 
algorithm known as POPQORN~\cite{ko2019popqorn}. The idea 
behind this technique is to observe the effect of outside disturbances to 
the model and thereby determine its robustness.

Of course, more and better data is always useful, and this is especially true
in free-text analysis. Having more realistic and longer-term free-text
datasets over a larger number of users would add credence to
the results obtained in any research.

\bibliographystyle{plain}
\bibliography{referencesHC.bib,Stamp-Mark.bib}

\end{document}

%% file: preambleHC.tex
\usepackage{amsmath,amsthm, amsfonts, amssymb, amsxtra,amsopn}
\usepackage{pgfplots}
\usepgfplotslibrary{colormaps}
\pgfplotsset{compat=1.15}
\usepackage{pgfplotstable}
\usetikzlibrary{pgfplots.statistics} 
\usepackage{filecontents}
\usepackage{graphicx}
\usepackage{multirow}
\usepackage{tabu}
\usepackage{algorithmic}
\usepackage{longtable}
\usepackage{booktabs}
\usepackage{listings}
\usepackage{cmap}
\usepackage{colortbl}
\usepackage{adjustbox}
\usepackage{epsfig}
\usepackage{xstring}

\usepackage{subfigure}

\PassOptionsToPackage{hyphens}{url}

\usepackage{hyperref}
\hypersetup{colorlinks=true,linkcolor=black,citecolor=black,urlcolor=blue,filecolor=black}
\hypersetup{pdfpagemode=UseNone,pdfstartview=}

\usepackage[tableposition=top]{caption}
\usepackage{enumitem}
\setlist[itemize]{noitemsep, topsep=0pt}

\advance\oddsidemargin by -0.45in
\advance\textwidth by 0.9in

\advance\topmargin by -0.375in
\advance\textheight by 0.75in

\long\def\symbolfootnotetext[#1]#2{\begingroup%
\def\thefootnote{\fnsymbol{footnote}}\footnotetext[#1]{#2}\endgroup}

\def\z{\phantom{0}}

\pgfkeys{
    /pgf/number format/fixed zerofill=true }
    
\pgfplotstableset{
    color cells/.code={%
        \pgfqkeys{/color cells}{#1}%
        \pgfkeysalso{%
            postproc cell content/.code={%
                \begingroup
                \pgfkeysgetvalue{/pgfplots/table/@preprocessed cell content}\value
\ifx\value\empty
\endgroup
\else
                \pgfmathfloatparsenumber{\value}%
                \pgfmathfloattofixed{\pgfmathresult}%
                \let\value=\pgfmathresult
                \pgfplotscolormapaccess
                    [\pgfkeysvalueof{/color cells/min}:\pgfkeysvalueof{/color cells/max}]%
                    {\value}%
                    {\pgfkeysvalueof{/pgfplots/colormap name}}%
                \pgfkeysgetvalue{/pgfplots/table/@cell content}\typesetvalue
                \pgfkeysgetvalue{/color cells/textcolor}\textcolorvalue
                \toks0=\expandafter{\typesetvalue}%
                \xdef\temp{%
                    \noexpand\pgfkeysalso{%
                        @cell content={%
                            \noexpand\cellcolor[rgb]{\pgfmathresult}%
                            \noexpand\definecolor{mapped color}{rgb}{\pgfmathresult}%
                            \ifx\textcolorvalue\empty
                            \else
                                \noexpand\color{\textcolorvalue}%
                            \fi
                            \the\toks0 %
                        }%
                    }%
                }%
                \endgroup
                \temp
\fi
            }%
        }%
    }
}

\def\srowvecc#1#2{(\!\begin{array}{cc} 
      \noexpandarg\IfBeginWith{#1}{-}{\! #1}{#1}
    & #2\kern-0.5pt\end{array}\!)}
\def\rowvecc#1#2{\left(\!\begin{array}{cc} 
      \noexpandarg\IfBeginWith{#1}{-}{\! #1}{#1}
    & #2\kern-0.5pt\end{array}\!\right)}
\def\rowveccc#1#2#3{\left(\!\begin{array}{ccc} 
      \noexpandarg\IfBeginWith{#1}{-}{\! #1}{#1}
    & #2 
    & #3\kern-0.5pt\end{array}\!\right)}
\def\rowvecccc#1#2#3#4{\left(\!\begin{array}{cccc}
      \noexpandarg\IfBeginWith{#1}{-}{\! #1}{#1}
    & #2 
    & #3 
    & #4\kern-0.5pt\end{array}\!\right)}
\def\srowvecccc#1#2#3#4{\bigl(\!\begin{array}{cccc}
      \noexpandarg\IfBeginWith{#1}{-}{\! #1}{#1}
    & #2 
    & #3 
    & #4\kern-0.5pt\end{array}\!\bigr)}
\def\rowveccccc#1#2#3#4#5{\left(\!\begin{array}{ccccc} 
      \noexpandarg\IfBeginWith{#1}{-}{\! #1}{#1}
    & #2
    & #3
    & #4
    & #5\kern-0.5pt\end{array}\!\right)}
\def\srowvecccccc#1#2#3#4#5#6{(\!\begin{array}{cccccc} 
      \noexpandarg\IfBeginWith{#1}{-}{\! #1}{#1}
    & #2
    & #3
    & #4
    & #5
    & #6\kern-0.5pt\end{array}\!)}
\def\rowvecccccc#1#2#3#4#5#6{\left(\!\begin{array}{cccccc} 
      \noexpandarg\IfBeginWith{#1}{-}{\! #1}{#1}
    & #2
    & #3
    & #4
    & #5
    & #6\kern-0.5pt\end{array}\!\right)}

\pgfkeys{
    /pgf/number format/precision=4, 
    /pgf/number format/fixed zerofill=true }
    
\pgfplotstableset{
    /color cells/min/.initial=0,
    /color cells/max/.initial=1000,
    /color cells/textcolor/.initial=,
    color cells/.code={%
        \pgfqkeys{/color cells}{#1}%
        \pgfkeysalso{%
            postproc cell content/.code={%
                \begingroup
                \pgfkeysgetvalue{/pgfplots/table/@preprocessed cell content}\value
\ifx\value\empty
\endgroup
\else
                \pgfmathfloatparsenumber{\value}%
                \pgfmathfloattofixed{\pgfmathresult}%
                \let\value=\pgfmathresult
                \pgfplotscolormapaccess
                    [\pgfkeysvalueof{/color cells/min}:\pgfkeysvalueof{/color cells/max}]%
                    {\value}%
                    {\pgfkeysvalueof{/pgfplots/colormap name}}%
                \pgfkeysgetvalue{/pgfplots/table/@cell content}\typesetvalue
                \pgfkeysgetvalue{/color cells/textcolor}\textcolorvalue
                \toks0=\expandafter{\typesetvalue}%
                \xdef\temp{%
                    \noexpand\pgfkeysalso{%
                        @cell content={%
                            \noexpand\cellcolor[rgb]{\pgfmathresult}%
                            \noexpand\definecolor{mapped color}{rgb}{\pgfmathresult}%
                            \ifx\textcolorvalue\empty
                            \else
                                \noexpand\color{\textcolorvalue}%
                            \fi
                            \the\toks0 %
                        }%
                    }%
                }%
                \endgroup
                \temp
\fi
            }%
        }%
    }
}


%% file: figures/barDiffUsers.tex
\begin{tikzpicture}[scale=0.95, every node/.style={scale=1.0}]
\begin{axis}[
        width  = 0.85*\textwidth,
        height = 7.5cm,
        ymin=0,ymax=125.0,
        ytick={0,20,40,60,80,100},
        major x tick style = transparent,
        ybar=5*\pgflinewidth,
        bar width=10.0pt,
        ylabel = {Accuracy (percentage)},
        symbolic x coords={User ID 1, User ID 2, User ID 3, User ID 4, User ID 5, User ID 6, User ID 7},
	y tick label style={
    		/pgf/number format/.cd,
   		fixed,
   		fixed zerofill,
    		precision=0},
        xtick = data,
        x tick label style={
        		rotate=60,
		font=\small,
		anchor=north east,
		inner sep=0mm
		},
        nodes near coords,
        every node near coord/.append style={rotate=90, 
        								   anchor=west,
								   font=\scriptsize,
								   /pgf/number format/.cd,
								   fixed,
								   fixed zerofill,
								   precision=2},
        enlarge x limits=0.085,
        legend cell align=left,
        legend style={
                at={(0.88,0.0175)},
                anchor=south,
                column sep=1ex
        },
]
\addplot [fill=blue,opacity=1.00]
coordinates {
(User ID 1, 64.28)
(User ID 2, 78.07)
(User ID 3, 81.88)
(User ID 4, 56.35)
(User ID 5, 92.34)
(User ID 6, 93.23)
(User ID 7, 91.86)
};
\addlegendentry{Timing}
\addplot [fill=orange,opacity=1.00]
coordinates {
(User ID 1, 81.25)
(User ID 2, 87.50)
(User ID 3, 81.25)
(User ID 4, 81.25)
(User ID 5, 93.75)
(User ID 6, 100.00)
(User ID 7, 87.50)
};
\addlegendentry{Rate}
\addplot [fill=green,opacity=1.00]
coordinates {
(User ID 1, 96.61)
(User ID 2, 100.00)
(User ID 3, 100.00)
(User ID 4, 70.34)
(User ID 5, 99.97)
(User ID 6, 94.15)
(User ID 7, 52.46)
};
\addlegendentry{Combined}
\end{axis}
\end{tikzpicture}

%% file: figures/smote1.tex
    \begin{tikzpicture}[scale=1.0]
 

    \node[circle,draw,thick,color=black,fill=gray,inner sep=1.75pt] (B) at  (8.0,4.3) {};
    \node[circle,draw,thick,color=black,fill=gray,inner sep=1.75pt] (C) at  (7.4,3.9) {};
    \node[circle,draw,thick,color=black,fill=gray,inner sep=1.75pt] (E) at  (7.9,3.8) {};
    \node[circle,draw,thick,color=black,fill=gray,inner sep=1.75pt] (F) at  (6.4,3.4) {};
    \node[circle,draw,thick,color=black,fill=gray,inner sep=1.75pt] (G) at  (8.35,4.9) {};
    \node[circle,draw,thick,color=black,fill=gray,inner sep=1.75pt] (H) at  (5.6,4.7) {};
    \node[circle,draw,thick,color=black,fill=gray,inner sep=1.75pt] (I) at  (7.1,3.0) {};
    \node[circle,draw,thick,color=black,fill=gray,inner sep=1.75pt] (J) at  (6.3,4.7) {};
    \node[circle,draw,thick,color=black,fill=gray,inner sep=1.75pt] (K) at  (7.45,4.45) {};
    \node[circle,draw,thick,color=black,fill=gray,inner sep=1.75pt] (L) at  (5.9,4.1) {};
    \node[circle,draw,thick,color=black,fill=gray,inner sep=1.75pt] (M) at  (7.275,3.5) {};
    \node[circle,draw,thick,color=black,fill=gray,inner sep=1.75pt] (N) at  (5.4,4.3) {};
    \node[circle,draw,thick,color=black,fill=gray,inner sep=1.75pt] (O) at  (5.6,3.75) {};
    \node[circle,draw,thick,color=black,fill=gray,inner sep=1.75pt] (P) at  (6.675,4.25) {};
    \node[circle,draw,thick,color=black,fill=gray,inner sep=1.75pt] (Q) at  (6.8,5.0) {};
    \node[circle,draw,thick,color=black,fill=gray,inner sep=1.75pt] (R) at  (6.25,3.8) {};
    \node[circle,draw,thick,color=black,fill=gray,inner sep=1.75pt] (S) at  (6.2,5.1) {};
    \node[circle,draw,thick,color=black,fill=gray,inner sep=1.75pt] (T) at  (7.4,5.0) {};

   
    \end{tikzpicture}

%% file: figures/smote3.tex
    \begin{tikzpicture}[scale=1.0]
 

    \node[circle,draw,thick,color=black,fill=gray,inner sep=1.75pt] (B) at  (8.0,4.3) {};
    \node[circle,draw,thick,color=black,fill=gray,inner sep=1.75pt] (C) at  (7.4,3.9) {};
    \node[circle,draw,thick,color=black,fill=gray,inner sep=1.75pt] (E) at  (7.9,3.8) {};
    \node[circle,draw,thick,color=black,fill=gray,inner sep=1.75pt] (F) at  (6.4,3.4) {};
    \node[circle,draw,thick,color=black,fill=gray,inner sep=1.75pt] (G) at  (8.35,4.9) {};
    \node[circle,draw,thick,color=black,fill=gray,inner sep=1.75pt] (H) at  (5.6,4.7) {};
    \node[circle,draw,thick,color=black,fill=gray,inner sep=1.75pt] (I) at  (7.1,3.0) {};
    \node[circle,draw,thick,color=black,fill=gray,inner sep=1.75pt] (J) at  (6.3,4.7) {};
    \node[circle,draw,thick,color=black,fill=gray,inner sep=1.75pt] (K) at  (7.45,4.45) {};
    \node[circle,draw,thick,color=black,fill=gray,inner sep=1.75pt] (L) at  (5.9,4.1) {};
    \node[circle,draw,thick,color=black,fill=gray,inner sep=1.75pt] (M) at  (7.275,3.5) {};
    \node[circle,draw,thick,color=black,fill=gray,inner sep=1.75pt] (N) at  (5.4,4.3) {};
    \node[circle,draw,thick,color=black,fill=gray,inner sep=1.75pt] (O) at  (5.6,3.75) {};
    \node[circle,draw,thick,color=black,fill=gray,inner sep=1.75pt] (P) at  (6.675,4.25) {};
    \node[circle,draw,thick,color=black,fill=gray,inner sep=1.75pt] (Q) at  (6.8,5.0) {};
    \node[circle,draw,thick,color=black,fill=gray,inner sep=1.75pt] (R) at  (6.25,3.8) {};
    \node[circle,draw,thick,color=black,fill=gray,inner sep=1.75pt] (S) at  (6.2,5.1) {};
    \node[circle,draw,thick,color=black,fill=gray,inner sep=1.75pt] (T) at  (7.4,5.0) {};

    \node[circle,draw,thick,color=red,inner sep=1.75pt] (a) at  (6.46,4.45) {};
    \draw[red] (J) -- (a);
    \draw[red] (P) -- (a);
    \node[circle,draw,thick,color=red,inner sep=1.75pt] (b) at  (5.9,4.9) {};
    \draw[red] (H) -- (b);
    \draw[red] (S) -- (b);
    \node[circle,draw,thick,color=red,inner sep=1.75pt] (c) at  (5.5,4.025) {};
    \draw[red] (N) -- (c);
    \draw[red] (O) -- (c);
    \node[circle,draw,thick,color=red,inner sep=1.75pt] (d) at  (6.75,3.2) {};
    \draw[red] (F) -- (d);
    \draw[red] (I) -- (d);
    \node[circle,draw,thick,color=red,inner sep=1.75pt] (e) at  (6.465,4.025) {};
    \draw[red] (R) -- (e);
    \draw[red] (P) -- (e);
    \node[circle,draw,thick,color=red,inner sep=1.75pt] (f) at  (6.825,3.45) {};
    \draw[red] (F) -- (f);
    \draw[red] (M) -- (f);
    \node[circle,draw,thick,color=red,inner sep=1.75pt] (g) at  (7.0,3.875) {};
    \draw[red] (M) -- (g);
    \draw[red] (P) -- (g);
    \node[circle,draw,thick,color=red,inner sep=1.75pt] (h) at  (6.725,4.625) {};
    \draw[red] (P) -- (h);
    \draw[red] (Q) -- (h);
    \node[circle,draw,thick,color=red,inner sep=1.75pt] (i) at  (6.1,4.4) {};
    \draw[red] (L) -- (i);
    \draw[red] (J) -- (i);

   
    \end{tikzpicture}

%% file: figures/lime.tex
\begin{tikzpicture}[scale=0.8]
    
 
    \draw[thick,blue] plot [smooth,tension=2] coordinates { (0,2) (1,3) (2,0) (3,2)};

    \draw[thick,color=black,fill=black] (1.0,3.5) circle (0.1);
    \draw[thick,color=black,fill=black] (1.5,3.2) circle (0.1);
    \draw[thick,color=black,fill=black] (1.65,2.7) circle (0.1);
    \draw[thick,color=black,fill=black] (1.75,2.0) circle (0.1);
    \draw[thick,color=black,fill=black] (2.0,1.5) circle (0.1);
    \draw[thick,color=black,fill=black] (2.15,1.1) circle (0.1);

    \draw[thick,color=red] (0.5,3.0) circle (0.1);
    \draw[thick,color=red] (1.0,2.5) circle (0.1);
    \draw[thick,color=red] (1.25,2.0) circle (0.1);
    \draw[thick,color=red] (0.35,2.65) circle (0.1);
    
    \draw[ultra thick,dashed] (0.15,4.0) -- (4.15,0.0);
    
    \draw (-0.125,0.0) -- (5.25,0.0);
    \draw (0.0,-0.125) -- (0.0,4.0);
    
%
%
  
\end{tikzpicture}

%% file: figures/barAccuracyPlusEER.tex
\begin{tikzpicture}[scale=0.95, every node/.style={scale=1.0}]
\begin{axis}[
xbar,
axis x line*=bottom,
width  = 0.65*\textwidth,
height = 8.75cm,
xmin=0,xmax=0.18,
xtick={0.00,0.05,0.10,0.15},
xlabel = {EER},
major y tick style = transparent,
xbar=5*\pgflinewidth,
bar width=9.0pt,
bar shift=-5.35pt,
x tick label style={
    	/pgf/number format/.cd,
   	fixed,
   	fixed zerofill,
    	precision=2},
y tick label style={
		font=\footnotesize,
		},
symbolic y coords={%
	CNN-GRU-without-sampler-fine-tune,
	CNN-GRU-with-sampler,
	CNN-GRU-without-sampler-not-best-val,
	CNN-GRU-without-sampler-best-EER,
	CNN-GRU-without-sampler-best-val,
	CNN-GRU-cross-entropy-loss,
	CNN-GRU-transformer-encoder,
	CNN-GRU-pre-trained-word-embedding},
ytick=data,
enlarge y limits=0.1,
nodes near coords, 
every node near coord/.append style={
	anchor=west,
	font=\footnotesize,
	/pgf/number format/.cd,
	fixed,
	fixed zerofill,
	precision=4},
ytick=data,
]
\addplot [fill=red,opacity=1.00]
coordinates {
(0.0386,CNN-GRU-without-sampler-fine-tune)
(0.0826,CNN-GRU-with-sampler)
(0.0594,CNN-GRU-without-sampler-not-best-val)
(0.0611,CNN-GRU-without-sampler-best-EER)
(0.0609,CNN-GRU-without-sampler-best-val)
(0.1502,CNN-GRU-cross-entropy-loss)
(0.1257,CNN-GRU-transformer-encoder)
(0.1091,CNN-GRU-pre-trained-word-embedding)
};
\label{EERplot}
\end{axis}
%
%
%
%
\begin{axis}[ 
xbar,
axis x line*=top,
width  = 0.65*\textwidth,
height = 8.75cm,
xmin=0,xmax=112.5,
xtick={0,20,40,60,80,100},
xlabel = {Accuracy (percentage)},
major y tick style = transparent,
xbar=5*\pgflinewidth,
bar width=9.0pt,
bar shift=5.35pt,
x tick label style={
    	/pgf/number format/.cd,
   	fixed,
   	fixed zerofill,
    	precision=0},
y tick label style={
		font=\footnotesize,
		},
symbolic y coords={%
	CNN-GRU-without-sampler-fine-tune,
	CNN-GRU-with-sampler,
	CNN-GRU-without-sampler-not-best-val,
	CNN-GRU-without-sampler-best-EER,
	CNN-GRU-without-sampler-best-val,
	CNN-GRU-cross-entropy-loss,
	CNN-GRU-transformer-encoder,
	CNN-GRU-pre-trained-word-embedding},
ytick=data,
enlarge y limits=0.1,
legend cell align=left,
legend style={
	at={(0.155,0.825)},
	anchor=south,
	column sep=1ex
},
nodes near coords, 
every node near coord/.append style={
	anchor=west,
	font=\footnotesize,
	/pgf/number format/.cd,
	fixed,
	fixed zerofill,
	precision=2},
ytick=data,
]
\addplot [fill=blue,opacity=1.00]
coordinates {
(99.06,CNN-GRU-without-sampler-fine-tune)
(98.24,CNN-GRU-with-sampler)
(98.20,CNN-GRU-without-sampler-not-best-val)
(97.84,CNN-GRU-without-sampler-best-EER)
(98.25,CNN-GRU-without-sampler-best-val)
(98.35,CNN-GRU-cross-entropy-loss)
(89.89,CNN-GRU-transformer-encoder)
(94.39,CNN-GRU-pre-trained-word-embedding)
};
\addlegendentry{Accuracy}
\addlegendimage{/pgfplots/refstyle=EERplot,color=black,fill=red}\addlegendentry{EER}
\end{axis}
\end{tikzpicture}

%% file: freeText_HC.bbl
\begin{thebibliography}{10}

\bibitem{9123909}
Blaine Ayotte, Mahesh Banavar, Daqing Hou, and Stephanie Schuckers.
\newblock Fast free-text authentication via instance-based keystroke dynamics.
\newblock {\em IEEE Transactions on Biometrics, Behavior, and Identity
  Science}, 2(4):377--387, 2020.

\bibitem{bib3}
Mario~Luca Bernardi, Marta Cimitile, Fabio Martinelli, and Francesco Mercaldo.
\newblock Keystroke analysis for user identification using deep neural
  networks.
\newblock In {\em 2019 International Joint Conference on Neural Networks},
  IJCNN, pages 1--8, 2019.

\bibitem{62613}
Saleh~Ali Bleha, Charles Slivinsky, and B.~Hussien.
\newblock Computer-access security systems using keystroke dynamics.
\newblock {\em IEEE Transactions on Pattern Analysis and Machine Intelligence},
  12(12):1217--1222, 1990.

\bibitem{10.1145/1150402.1150464}
Cristian Bucilu\u{a}, Rich Caruana, and Alexandru Niculescu-Mizil.
\newblock Model compression.
\newblock In {\em Proceedings of the 12th ACM SIGKDD International Conference
  on Knowledge Discovery and Data Mining}, KDD '06, pages 535--541, 2006.

\bibitem{DBLP:journals/corr/abs-1810-04805}
Jacob Devlin, Ming{-}Wei Chang, Kenton Lee, and Kristina Toutanova.
\newblock {BERT:} pre-training of deep bidirectional transformers for language
  understanding.
\newblock \url{http://arxiv.org/abs/1810.04805}, 2018.

\bibitem{10.1007/1-4020-8143-X_18}
Paul~S. Dowland and Steven~M. Furnell.
\newblock A long-term trial of keystroke profiling using digraph, trigraph and
  keyword latencies.
\newblock In Yves Deswarte, Fr{\'e}d{\'e}ric Cuppens, Sushil Jajodia, and
  Lingyu Wang, editors, {\em Security and Protection in Information Processing
  Systems}, pages 275--289. Springer, 2004.

\bibitem{forsen1977personal}
George~E. Forsen, Mark~R. Nelson, and Raymond J.~Staron (Jr.).
\newblock Personal attributes authentication techniques.
\newblock \url{https://books.google.com.tw/books?id=tbs4OAAACAAJ}, 1977.

\bibitem{ger2020autoencoders}
Stephanie Ger and Diego Klabjan.
\newblock Autoencoders and generative adversarial networks for imbalanced
  sequence classification.
\newblock \url{https://arxiv.org/abs/1901.02514}, 2020.

\bibitem{Siamese}
Romain Giot and Anderson Rocha.
\newblock Siamese networks for static keystroke dynamics authentication.
\newblock In {\em 2019 IEEE International Workshop on Information Forensics and
  Security}, WIFS, pages 1--6, 2019.

\bibitem{886039}
S.~{Haider}, A.~{Abbas}, and A.~K. {Zaidi}.
\newblock A multi-technique approach for user identification through keystroke
  dynamics.
\newblock In {\em 2000 IEEE International Conference on Systems, Man and
  Cybernetics}, SMC 2000, pages 1336--1341, 2000.

\bibitem{cuda}
Martin Heller.
\newblock \textit{InfoWorld}: What is {CUDA}? {P}arallel programming for
  {GPU}s.
\newblock
  \url{https://www.infoworld.com/article/3299703/what-is-cuda-parallel-programming-for-gpus.html},
  2018.

\bibitem{hinton2015distilling}
Geoffrey Hinton, Oriol Vinyals, and Jeff Dean.
\newblock Distilling the knowledge in a neural network.
\newblock \url{https://arxiv.org/abs/1503.02531}, 2015.

\bibitem{4656564}
Danoush Hosseinzadeh and Sridhar Krishnan.
\newblock Gaussian mixture modeling of keystroke patterns for biometric
  applications.
\newblock {\em IEEE Transactions on Systems, Man, and Cybernetics, Part C
  (Applications and Reviews)}, 38(6):816--826, 2008.

\bibitem{8267670}
Jiaju Huang, Daqing Hou, Stephanie Schuckers, Timothy Law, and Adam Sherwin.
\newblock Benchmarking keystroke authentication algorithms.
\newblock In {\em 2017 IEEE Workshop on Information Forensics and Security},
  WIFS, pages 1--6, 2017.

\bibitem{10.1007/978-3-540-74549-5_125}
Pilsung Kang, Seong-seob Hwang, and Sungzoon Cho.
\newblock Continual retraining of keystroke dynamics based authenticator.
\newblock In Seong-Whan Lee and Stan~Z. Li, editors, {\em Advances in
  Biometrics}, pages 1203--1211. Springer, 2007.

\bibitem{ko2019popqorn}
Ching-Yun Ko, Zhaoyang Lyu, Tsui-Wei Weng, Luca Daniel, Ngai Wong, and Dahua
  Lin.
\newblock {POPQORN}: Quantifying robustness of recurrent neural networks.
\newblock In {\em Proceedings of the 36th International Conference on Machine
  Learning}, pages 3468--3477, 2019.

\bibitem{10}
Xiaofeng Lu, Shengfei Zhang, and Shengwei Yi.
\newblock Free-text keystroke continuous authentication using {CNN} and {RNN}.
\newblock {\em Journal of Tsinghua University (Science and Technology)},
  58(12):1072--1078, 2018.

\bibitem{5}
Roy~A. {Maxion} and Kevin~S. {Killourhy}.
\newblock Comparing anomaly-detection algorithms for keystroke dynamics.
\newblock In {\em 2009 IEEE/IFIP International Conference on Dependable Systems
  \&\ Networks}, DSN, pages 125--134, 2009.

\bibitem{5544311}
Roy~A. {Maxion} and Kevin~S. {Killourhy}.
\newblock Keystroke biometrics with number-pad input.
\newblock In {\em 2010 IEEE/IFIP International Conference on Dependable Systems
  \&\ Networks}, DSN, pages 201--210, 2010.

\bibitem{w2v1}
Tomas Mikolov, Kai Chen, Greg Corrado, and Jeffrey Dean.
\newblock \url{https://arxiv.org/ abs/1301.3781}, 2013.

\bibitem{w2v2}
Tomas Mikolov, Ilya Sutskever, Kai Chen, Greg Corrado, and Jeffrey Dean.
\newblock Distributed representations of words and phrases and their
  compositionality.
\newblock \url{https://arxiv.org/abs/1310.4546}, 2013.

\bibitem{Monrose}
Fabian Monrose and Aviel Rubin.
\newblock Authentication via keystroke dynamics.
\newblock In {\em Proceedings of the 4th ACM Conference on Computer and
  Communications Security}, pages 48--56, 1997.

\bibitem{MONROSE2000351}
Fabian Monrose and Aviel~D. Rubin.
\newblock Keystroke dynamics as a biometric for authentication.
\newblock {\em Future Generation Computer Systems}, 16(4):351--359, 2000.

\bibitem{NEURIPS2019_9015}
Adam Paszke, Sam Gross, Francisco Massa, Adam Lerer, James Bradbury, Gregory
  Chanan, Trevor Killeen, Zeming Lin, Natalia Gimelshein, Luca Antiga, Alban
  Desmaison, Andreas Kopf, Edward Yang, Zachary DeVito, Martin Raison, Alykhan
  Tejani, Sasank Chilamkurthy, Benoit Steiner, Lu~Fang, Junjie Bai, and Soumith
  Chintala.
\newblock {PyTorch}: An imperative style, high-performance deep learning
  library.
\newblock In H.~Wallach, H.~Larochelle, A.~Beygelzimer, F.~d'~Alch\'{e}-Buc,
  E.~Fox, and R.~Garnett, editors, {\em Advances in Neural Information
  Processing Systems 32}, pages 8024--8035. Curran Associates, Inc., 2019.

\bibitem{4}
Nataasha Raul, Radha Shankarmani, and Padmaja Joshi.
\newblock A comprehensive review of keystroke dynamics-based authentication
  mechanism.
\newblock In {\em International Conference on Innovative Computing and
  Communications}, pages 149--162, 2020.

\bibitem{lime}
Marco~Tulio Ribeiro, Sameer Singh, and Carlos Guestrin.
\newblock Why should {I} trust you?: Explaining the predictions of any
  classifier.
\newblock In {\em Proceedings of the 22nd {ACM} International Conference on
  Knowledge Discovery and Data Mining}, SIGKDD, pages 1135--1144, 2016.

\bibitem{6966780}
Joseph Roth, Xiaoming Liu, Arun Ross, and Dimitris Metaxas.
\newblock Investigating the discriminative power of keystroke sound.
\newblock {\em IEEE Transactions on Information Forensics and Security},
  10(2):333--345, 2015.

\bibitem{4270391}
T.~{Sim} and R.~{Janakiraman}.
\newblock Are digraphs good for free-text keystroke dynamics?
\newblock In {\em 2007 IEEE Conference on Computer Vision and Pattern
  Recognition}, pages 1--6, 2007.

\bibitem{Srivastava}
Nitish Srivastava, Geoffrey Hinton, Alex Krizhevsky, Ilya Sutskever, and Ruslan
  Salakhutdinov.
\newblock Dropout: A simple way to prevent neural networks from overfitting.
\newblock {\em Journal of Machine Learning Research}, 15:1929--1958, 2014.

\bibitem{9}
Yan Sun, Hayreddin \c{C}eker, and Shambhu Upadhyaya.
\newblock Shared keystroke dataset for continuous authentication.
\newblock In {\em 2016 IEEE International Workshop on Information Forensics and
  Security}, WIFS, pages 1--6, 2016.

\bibitem{122}
Pin~Shen Teh, Andrew Teoh, and Shigang Yue.
\newblock A survey of keystroke dynamics biometrics.
\newblock {\em The Scientific World Journal}, 2013, 2013.
\newblock Article ID 408280.

\bibitem{5634492}
Robert~S. Zack, Charles~C. Tappert, and Sung-Hyuk Cha.
\newblock Performance of a long-text-input keystroke biometric authentication
  system using an improved $k$-nearest-neighbor classification method.
\newblock In {\em 2010 Fourth IEEE International Conference on Biometrics:
  Theory, Applications and Systems}, BTAS, pages 1--6, 2010.

\bibitem{Zhong}
Yu~Zhong and Yunbin Deng.
\newblock A survey on keystroke dynamics biometrics: Approaches, advances, and
  evaluations.
\newblock
  \url{https://sciencegatepub.com/books/gcsr/gcsr_vol2/GCSR_Vol2_Ch1.pdf},
  2015.

\bibitem{11}
Yu~Zhong, Yunbin Deng, and Anil~K. Jain.
\newblock Keystroke dynamics for user authentication.
\newblock In {\em 2012 IEEE Computer Society Conference on Computer Vision and
  Pattern Recognition Workshops}, pages 117--123, 2012.

\end{thebibliography}
